\documentclass[letterpaper, 10 pt, conference]{IEEE/ieeeconf}

\ifdefined\pdfobjcompresslevel\pdfobjcompresslevel=0\fi

\IEEEoverridecommandlockouts 

\usepackage{amsmath} %
\usepackage{amssymb}  %
\usepackage{multirow} %
\usepackage{array} %
\usepackage{graphicx} %
\usepackage{siunitx} %
\usepackage{bm} %
\usepackage{xcolor}
\usepackage{capt-of}
\usepackage[font=footnotesize]{caption}
\let\labelindent\relax
\usepackage{enumitem}
\usepackage{cite} %

\title{\LARGE \bf
Occupancy Planar Patch: Incremental Probabilistic Bounded-Plane Reconstruction from LiDAR Data
}

\author{Jiahao Wang, Chaehyeon Song, Dongjiao He, Jan Quenzel, Lintong Zhang, Ayoung Kim and Maurice Fallon}

\IEEEaftertitletext{
        \noindent
        \begin{minipage}{\textwidth}
                \centering
                \newlength{\herow}\setlength{\herow}{0.2415\textwidth}
                \setlength{\tabcolsep}{1pt}
                \renewcommand{\arraystretch}{1.1}
                \begin{tabular}{@{}cccc@{}}
                        \parbox[b]{\herow}{\centering\footnotesize Tripod Scan (Not Used)}
                        & \parbox[b]{\herow}{\centering\footnotesize Volumetric Mesh (VDBFusion)}
                        & \parbox[b]{\herow}{\centering\footnotesize OPP Model (Ours)}
                        & \parbox[b]{\herow}{\centering\footnotesize Textured OPP Model (Ours)} \\[2pt]
                        \includegraphics[width=\herow]{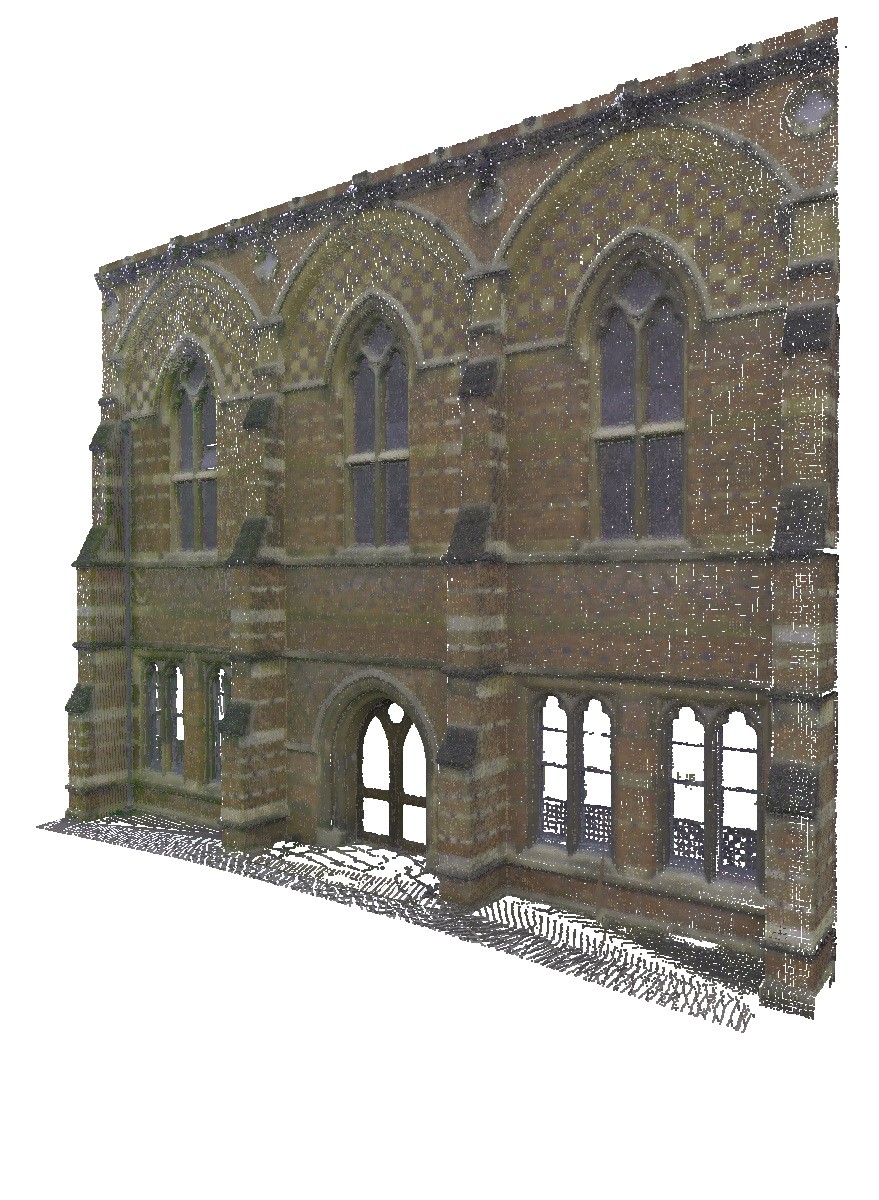}
                        & \includegraphics[width=\herow]{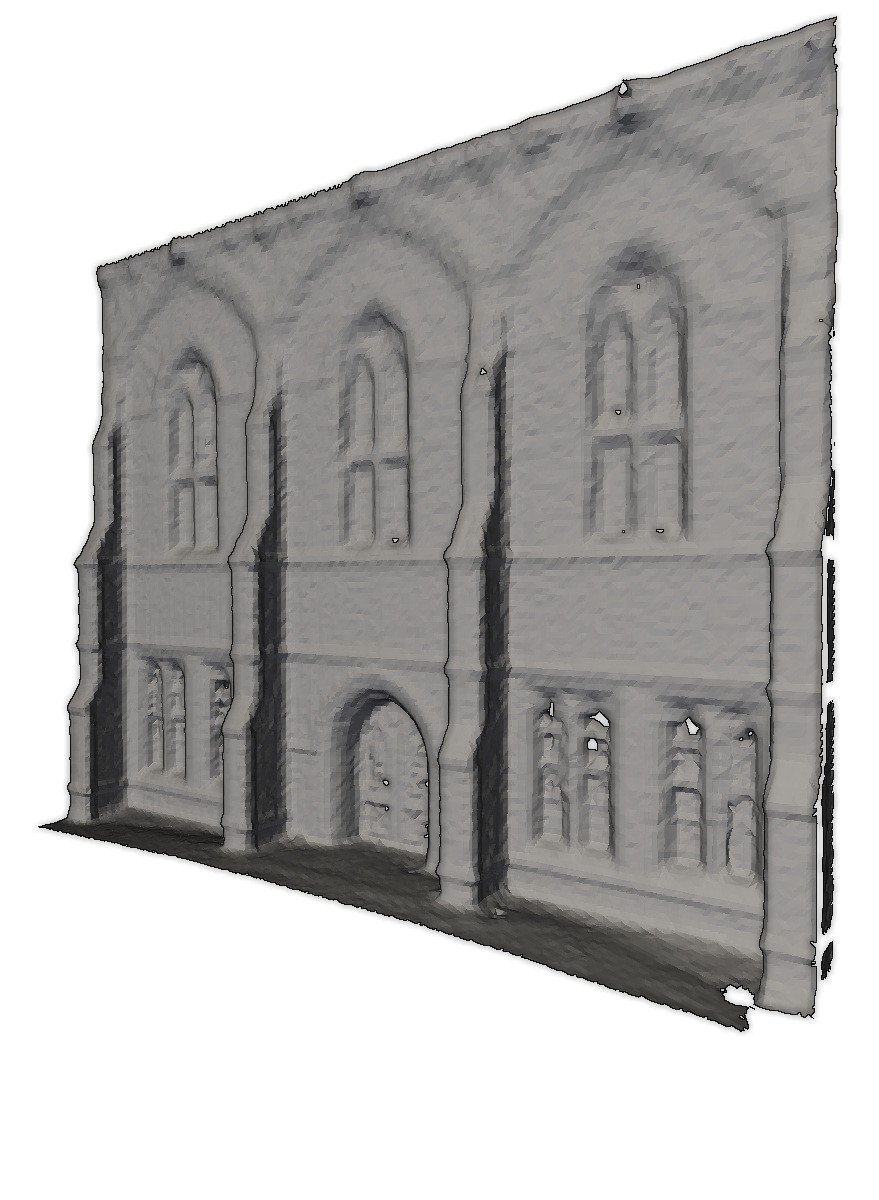}
                        & \includegraphics[width=\herow]{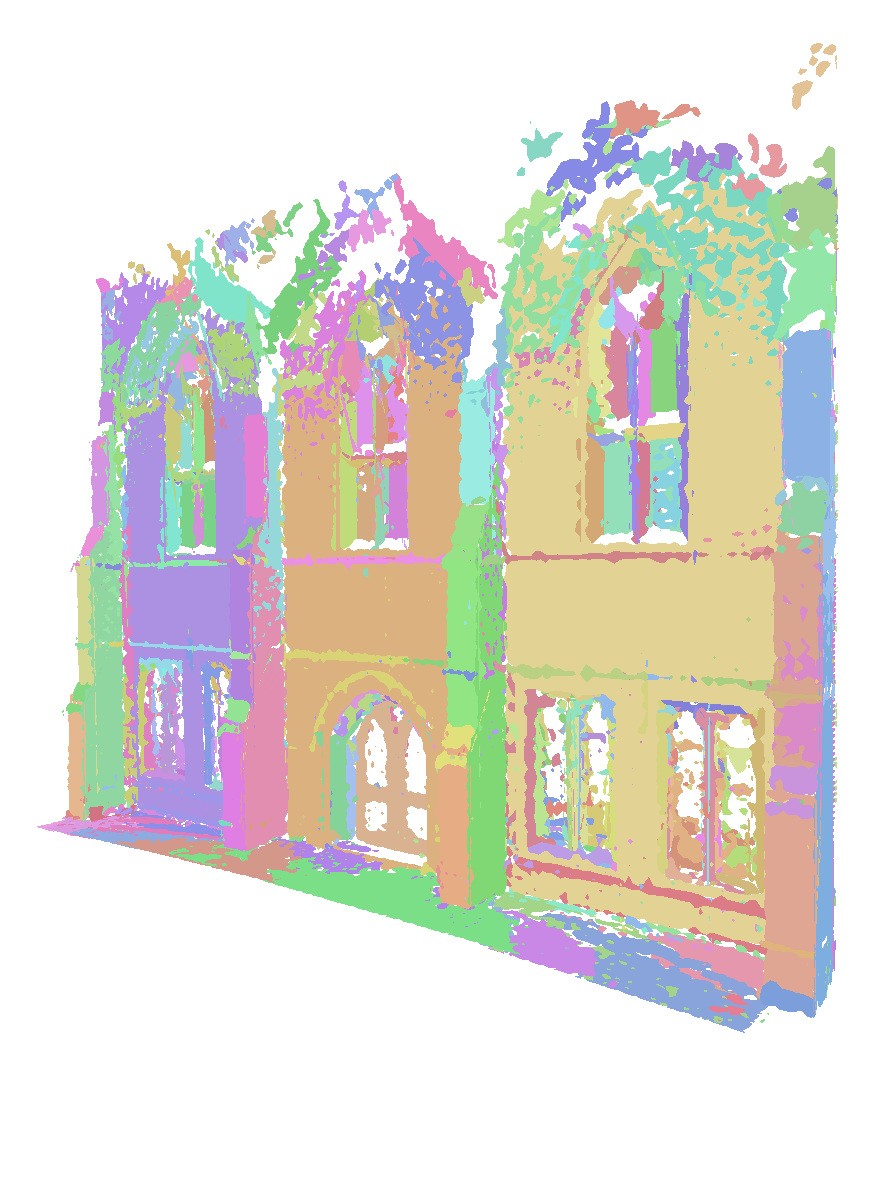}
                        & \includegraphics[width=\herow]{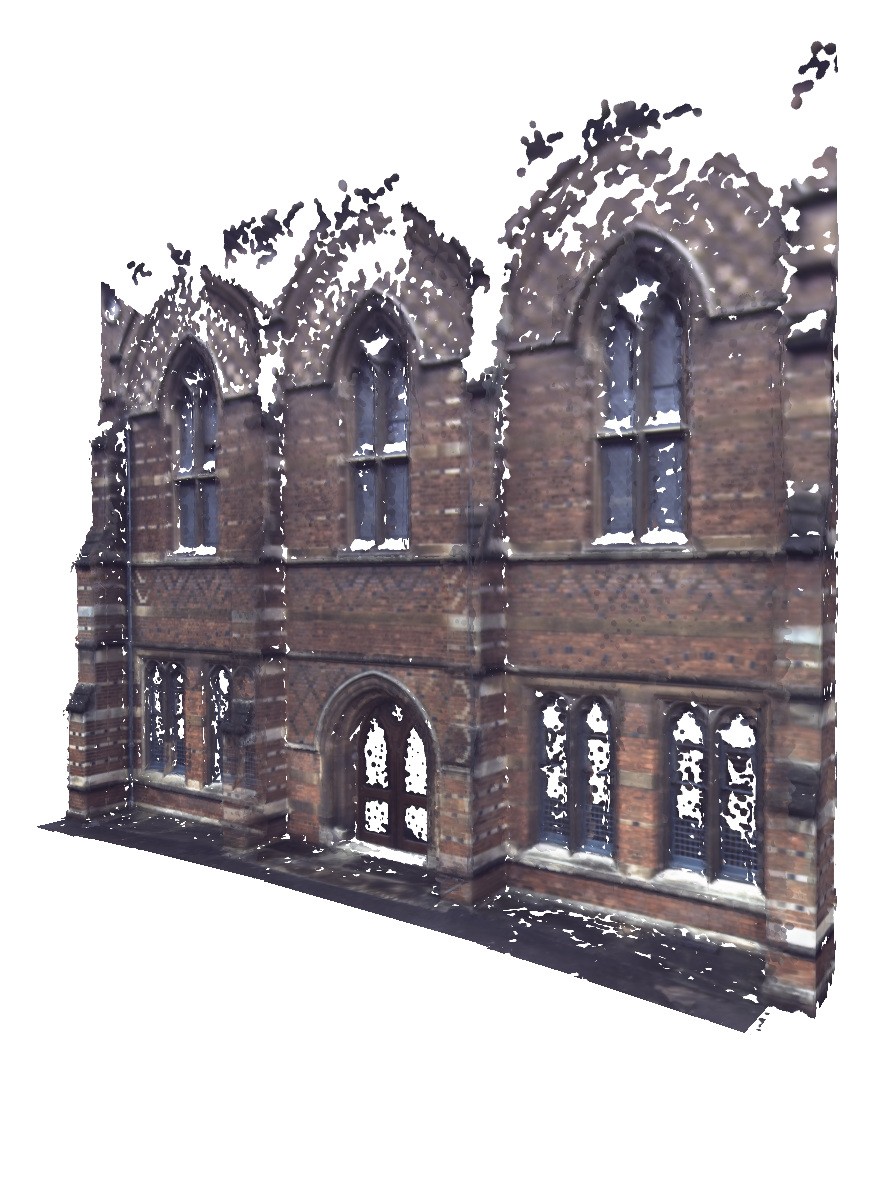} \\[-8.5ex]
                        {\footnotesize (a)} & {\footnotesize (b)} & {\footnotesize (c)} & {\footnotesize (d)} \\
                \end{tabular}
                \captionof{figure}{Online reconstruction with Occupancy Planar Patches (OPP) on a building facade from the Oxford Spires Dataset (OSD)~\cite{Oxford-Spires-Dataset}. The views show: (a) ground-truth tripod reference scan (not used during reconstruction); (b) volumetric mesh (VDBFusion~\cite{VDBFusion}), which oversmooths angular window reveals and brickwork into the surrounding wall; (c) OPP reconstruction with uniquely coloured planar patches, preserving depth discontinuities and maintaining structural elements as distinct surfaces; and (d) the OPP model rendered with texture.}
                \label{fig:banner_top}
        \end{minipage}
        \vspace{1em}
}

\newcommand{\pose}{\mathbf{T}}                 %
\newcommand{\cloud}{\mathcal{P}}               %
\newcommand{\image}{\mathcal{I}}               %
\newcommand{\origin}{\mathbf{o}}               %
\newcommand{\direction}{\mathbf{d}}            %
\newcommand{\range}{r}                         %
\newcommand{\expectedrange}{\bar{r}}           %
\newcommand{\patchset}{\mathcal{K}}            %
\newcommand{\plane}{\bm{\pi}}                  %
\newcommand{\normal}{\mathbf{n}}               %
\newcommand{\distance}{d}                      %
\newcommand{\dloc}{\distance_{\mathrm{local}}}  %
\newcommand{\pivot}{\mathbf{c}}                %
\newcommand{\hit}{\mathbf{p}}                  %
\newcommand{\uv}{\mathbf{u}}                   %
\newcommand{\basis}{\mathbf{E}}                %
\newcommand{\occ}{O}                           %
\newcommand{\kernel}{\psi}                     %
\newcommand{\logodds}{\phi}                    %
\newcommand{\krad}{\rho}                       %
\newcommand{\logistic}{\varsigma}              %
\newcommand{\amp}{\alpha}                      %
\newcommand{\atlas}{A}                         %
\newcommand{\patchcov}[1][]{\Sigma_{\plane#1}} %
\newcommand{\npatch}{M}                        %
\newcommand{\nkernel}{K}                       %
\newcommand{\nazim}{H}                         %
\newcommand{\nelev}{V}                         %
\newcommand{\resp}{\gamma}                     %
\newcommand{\respmin}{\resp_{\min}}            %
\newcommand{\temp}{\tau}                       %
\newcommand{\dtheta}{\delta\bm{\theta}}        %
\newcommand{\roughness}{\mathrm{rough}}        %
\newcommand{\cdf}{\Phi}                        %
\newcommand{\pin}[1]{P_{\mathrm{in},#1}}       %
\newcommand{\pout}[1]{P_{\mathrm{out},#1}}     %
\newcommand{\ptrans}{P_{\mathrm{trans}}}       %
\newcommand{\pexcl}{P_{\mathrm{excl}}}         %
\newcommand{\pvis}{P_{\mathrm{vis}}}           %
\newcommand{\reffig}[1]{Fig.~\ref{#1}}
\newcommand{\reftab}[1]{Table~\ref{#1}}
\newcommand{\refsec}[1]{Sec.~\ref{#1}}
\newcommand{\refeq}[1]{Eq.~\eqref{#1}}

\let\oldthebibliography\thebibliography
\renewcommand{\thebibliography}[1]{%
  \oldthebibliography{#1}%
  \setlength{\itemsep}{-1pt}%
  \setlength{\parskip}{0pt}%
}

\begin{document}

\maketitle
\thispagestyle{empty}
\pagestyle{empty}

\begin{abstract}
        Online surface reconstruction from LiDAR scans requires observation models that respect the sensor's physical measurement geometry. Unlike unordered point sets, LiDAR returns carry anisotropic range noise along known rays, which also implicitly identify free space.
        We introduce the Occupancy Planar Patch (OPP), an incremental bounded planar patch representation that evaluates surface likelihoods and free space directly in range-space, allowing sensor noise, plane uncertainty, and ray clearance to be explicitly modelled during reconstruction. A single ray-to-patch association weight jointly updates plane geometry, refines surface boundaries, and blends camera imagery.
        The method is evaluated using six sequences from the Oxford Spires and Multi-Campus datasets. OPP achieves the highest precision and lowest reconstruction error among evaluated methods, while producing stored maps among the most compact. By conservatively admitting surfaces only where geometric evidence is certain, OPP prioritises structural precision over unconstrained completeness, which is well suited for architectural modelling. Geometric reconstruction operates in real time, requiring \SI{81}{\milli\second} per frame on a representative sequence, with texture projection adding \SI{134}{\milli\second}.
\end{abstract}

\section{INTRODUCTION}        
        Mobile robots and handheld LiDAR scanners can map a large building in a single walk-through~\cite{Oxford-Spires-Dataset, FAST-LIO2}. However, raw point clouds lack explicit representations of solid surfaces and free space. Recovering explicit surfaces is necessary to make the data usable downstream: for digital twinning~\cite{Scan-to-BIM-Review}, construction monitoring~\cite{Bosche2010}, simulation, motion planning~\cite{Mesh-Navigation}, and localisation~\cite{Puma}.
        However, downstream applications impose differing requirements on surface reconstructions. Physics simulation and collision checking typically require a single watertight mesh. By contrast, for architectural modelling, surfaces should be separate and individually parameterised so that walls, window reveals, and structural elements remain distinct objects~\cite{Scan-to-BIM-Review}. In this work, we focus on architectural modelling and perform surface reconstruction in real time during scanning.

        Conventional surface reconstruction often treats scans as unordered point sets with isotropic noise~\cite{Puma, PolyFit}. A LiDAR scan, however, is a collection of range measurements from the sensor origin along known rays, with noise acting predominantly along those rays and the space in front of each return known to be empty~\cite{Probabilistic-robotics, A-volumetric-method-for-building-complex-models-from-range-images}. As a result, we formulate our method directly in range-space to exploit the sensor measurement geometry, preserving directional and free-space cues without voxelising 3D space.

        Building interiors, facades, corridors, and urban streets are predominantly piecewise planar~\cite{Manhattan-World, Piecewise-Planar-Urban-Stereo}, making bounded planar patches an effective map representation. Unlike local volumetric voxels~\cite{OctoMap, VDBFusion}, a plane pools range measurements over its entire physical support, continuously refining the estimate as the surface is observed from different viewpoints during a walk-through. Planar parameterisations can also represent large structures with minimal storage~\cite{PlanarMesh} and directly provide the geometric primitives required by downstream architectural and CAD pipelines~\cite{Scan-to-BIM-Review}.

        We introduce the Occupancy Planar Patch (OPP), an incremental mapping framework designed to reconstruct clean, structured geometry in real time (\reffig{fig:banner_top}). OPP models architectural surfaces as supporting planes whose spatial extents are continuously refined by sensor visibility and free-space carving. By formulating inference directly in the LiDAR's measurement space, the reconstructed model evolves recursively with each incoming beam. 
        
        \noindent In summary, our contributions are:
        \begin{itemize}[leftmargin=*, noitemsep, topsep=2pt]
                \item \textbf{Occupancy Planar Patch:} a bounded planar patch representation coupling an infinite supporting plane with a continuous 2D occupancy field whose boundary is delineated by its iso-contour, preserving depth discontinuities and maintaining structural elements as distinct surfaces.
                \item \textbf{Unified recursive inference in range-space:} evaluating predictive likelihoods directly in range-space models anisotropic noise and uncertainties in closed form; a single per-ray responsibility jointly drives recursive plane refinement, continuous boundary learning, and texture projection while maintaining a valid map after every scan.
        \end{itemize}

\section{RELATED WORK}        
        Incremental reconstruction from LiDAR belongs to dense mapping and simultaneous localisation and mapping (SLAM), where the map is maintained in a low-level representation rather than as raw points. Common representations include volumetric occupancy~\cite{OctoMap}, signed distance fields~\cite{VDBFusion, Kimera, PINGS}, surfels~\cite{SuMa}, edge and plane features~\cite{LOAM}, and meshes, either extracted offline~\cite{R3LIVE} or maintained online~\cite{ImMesh}. Conceptually closest to our approach is the normal distributions transform~\cite{NDT, NDT3D}, which also models the scene as a Gaussian mixture; it places one Gaussian per cell of a fixed grid describing point distributions, whereas we assign one component per planar surface and evaluate it along the ray, directly incorporating the sensor noise model and ray clearance.
        Because built environments are predominantly planar, planar representations are widely used, either as the primary map representation~\cite{Planar-Surface-SLAM} or as geometric features for odometry~\cite{LOAM}. Existing formulations differ in spatial extent: unbounded infinite planes~\cite{Simultaneous-Localization-and-Mapping-with-Infinite-Planes}, voxel-bounded planes~\cite{BALM}, or explicitly bounded planar segments~\cite{Planar-Surface-SLAM, 3D-SLAM-using-planar-segments}. We build on bounded representations. The following subsections review the three core decisions in bounded-plane mapping: data association, plane and boundary parameterisation, and appearance integration.

        \subsection{Data Association}
                Existing planar mapping pipelines predominantly associate measurements via RANSAC~\cite{RANSAC, PolyFit}, range-image region growing~\cite{RegionGrowing}, or voxel subdivision~\cite{BALM}. While computationally simple, RANSAC fits infinite planes and fails to separate disconnected coplanar surfaces; region growing assumes pixel adjacency and breaks at depth discontinuities; and voxelisation imposes an arbitrary grid scale where cell size dictates boundary resolution.
                Recent bounded-plane frameworks perform per-ray association against candidate surfaces~\cite{PlanarMesh}, but hard assignment forces premature decisions when multiple surfaces explain an observation with comparable likelihood. We adopt soft assignment inspired by mixture models~\cite{MARS, The-infinite-Gaussian-mixture-model}, evaluating candidate responsibilities directly along each ray so that sensor uncertainty and plane geometry dictate association confidence.

        \subsection{Plane Fitting and Boundaries}\label{sec:boundaries}
                Plane parameters are conventionally extracted via PCA~\cite{PolyFit} or planar bundle adjustment~\cite{BALM}. Both minimise Euclidean point-to-plane distances under an isotropic noise assumption, introducing systematic bias when applied to the anisotropic range noise of LiDAR sensors. Evaluating residuals directly in range-space along measurement rays eliminates this bias.
                For spatial extent, previous methods rely on explicit boundaries such as convex hulls~\cite{Planar-Surface-SLAM}, polygons~\cite{PlanarMesh}, or alpha shapes~\cite{3D-SLAM-using-planar-segments}. These binary representations force hard inside/outside decisions regardless of observation density. In contrast, an implicit continuous field retains partial evidence and naturally accommodates sparsely sampled regions.

        \subsection{Appearance}
                Downstream robotic tasks such as visual localisation, simulation, and semantic mapping benefit from geometry carrying appearance or semantics~\cite{Kimera}. In real-time dense mapping, colour is typically fused via running averages~\cite{Point-based-fusion} or radial-distance weights~\cite{ElasticFusion}. Surface visibility is conventionally resolved by Z-buffering to select the closest surface, whereas the A-buffer~\cite{The-A-buffer-an-antialiased-hidden-surface-method} maintains a depth-sorted list of intersections per ray to allow multi-layer depth composition.

\section{METHOD}        
\begin{figure*}[ht]
                \centering
                \includegraphics[width=\textwidth]{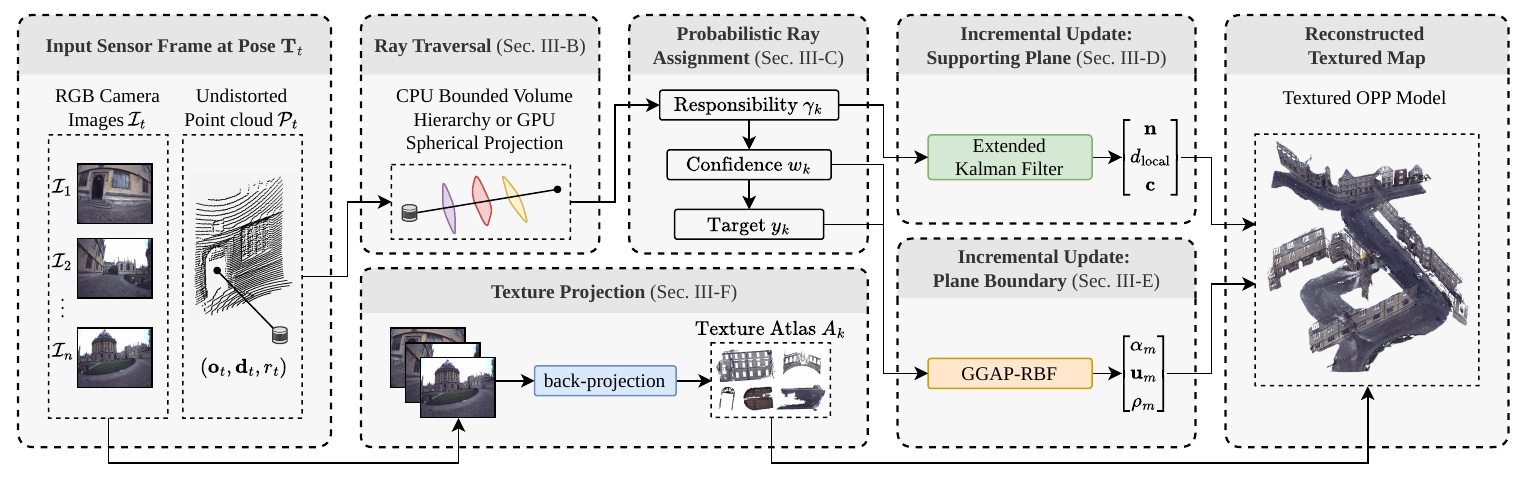}
                \caption{One integration step of the OPP reconstruction framework. Given sensor poses $\pose_t$, LiDAR point clouds $\cloud_t$, and RGB images $\image_t$, each motion-undistorted LiDAR ray $(\origin_t, \direction_t, \range_t)$ is traced to query intersecting candidate patches via spatial acceleration (\refsec{sec:search}), evaluated in range-space to compute soft assignment responsibilities $\resp_k$ (\refsec{sec:assignment}), and used to concurrently update the supporting plane via an extended Kalman filter (\refsec{sec:geometry}) and its continuous 2D occupancy boundary via GGAP-RBF~\cite{GGAP-RBF} (\refsec{sec:occupancy}). Finally, visual appearance from camera images is projected onto texture atlases $\atlas_k$ to update the textured OPP model (\refsec{sec:texture}).}
                \label{fig:overview}
        \end{figure*}

        We present the OPP reconstruction framework, which incrementally processes LiDAR point clouds $\cloud_t$, RGB images $\image_t$, and sensor poses $\pose_t \in \mathrm{SE}(3)$ to reconstruct a collection of textured planar patches $\patchset$. As illustrated in \reffig{fig:overview}, each LiDAR return is motion-undistorted and traced as a ray to query intersecting candidate patches (\refsec{sec:search}). It is then evaluated in range-space to compute a soft ray-to-patch assignment weight (\refsec{sec:assignment}), and used to concurrently update the supporting plane (\refsec{sec:geometry}) and its continuous occupancy boundary (\refsec{sec:occupancy}). Finally, incoming images are projected incrementally onto the texture atlas (\refsec{sec:texture}).

        \subsection{Scene Representation: The Occupancy Planar Patch}\label{sec:representation}
\begin{figure}[ht]
                        \centering
                        \newlength{\anatw}\setlength{\anatw}{0.32\columnwidth}
                        \setlength{\tabcolsep}{1pt}
                        \begin{tabular}{@{}c@{}c@{}c@{}}
                                \includegraphics[width=\anatw]{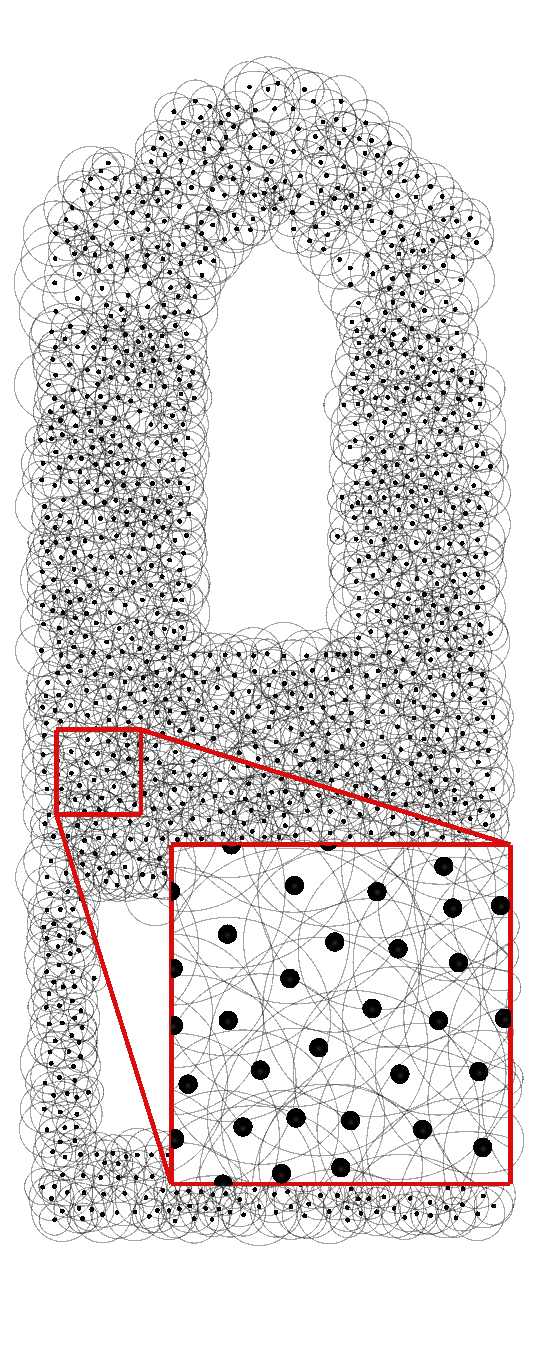}
                                & \includegraphics[width=\anatw]{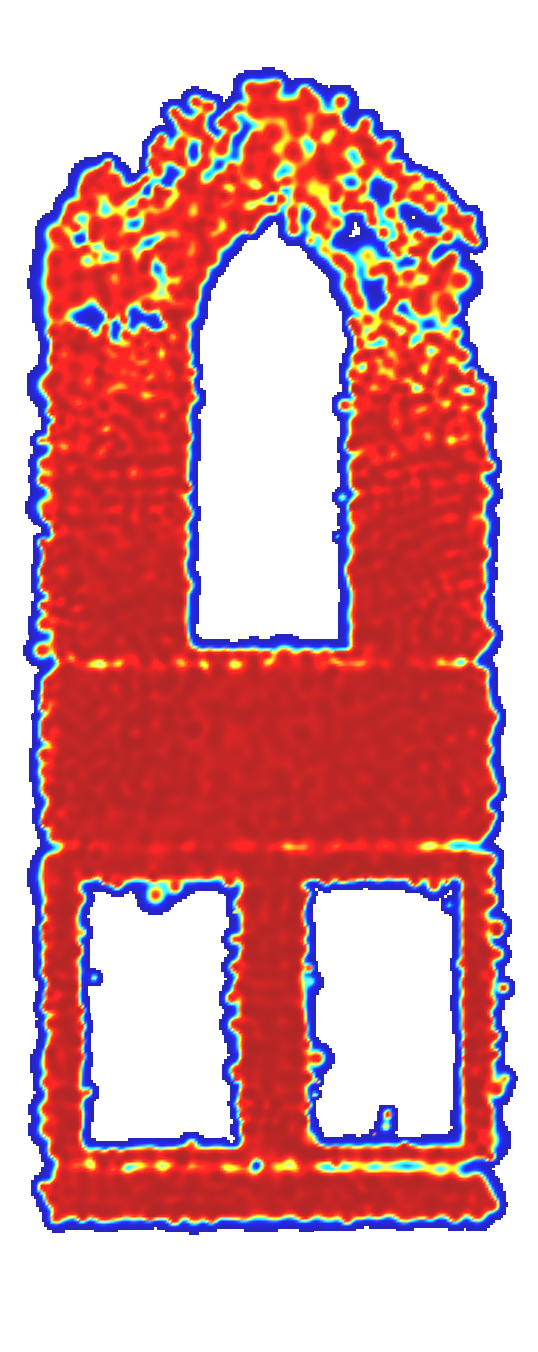}
                                & \includegraphics[width=\anatw]{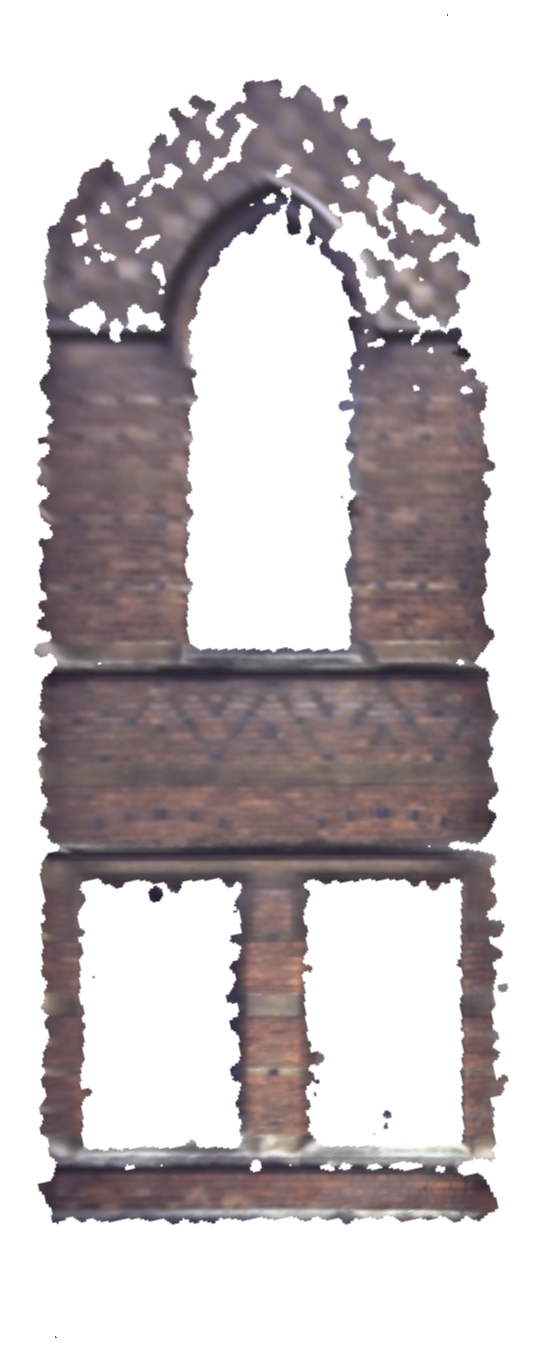} \\[-4.5ex]
                                {\footnotesize (a)} & {\footnotesize (b)} & {\footnotesize (c)} \\
                        \end{tabular}
                        \caption{Continuous boundary and texture layers are defined on a patch's supporting plane: (a) allocated CSRBF kernels with their centres and radii; (b) the resulting 2D occupancy field (red is occupied, blue is free space); and (c) the associated texture atlas.}
                        \label{fig:patch_anatomy}
                \end{figure}

                The scene is represented as a collection of textured planar patches $\patchset = \{(\plane_k, \occ_k, \atlas_k)\}$. Each patch $k$ couples an infinite supporting plane $\plane_k$ with a continuous occupancy field $\occ_k$ defining its finite spatial boundary, alongside a texture atlas $\atlas_k$ capturing surface appearance (\reffig{fig:patch_anatomy}). This formulation mirrors planar surfaces in real-world architecture, which are physically flat across their span yet strictly bounded by corners, edges, and openings.

                \subsubsection{Supporting Plane Parameterisation}
                The supporting plane comprises a unit normal $\normal$, an orthonormal basis $\basis = [\mathbf{e}_1, \mathbf{e}_2]$ of its tangent space, and a distance $\dloc$ measured from a pivot point $\pivot$, the running weighted average of all assigned points.
                A rotation vector $\dtheta \in \mathbb{R}^3$ perturbs the normal $\normal$ as $\exp([\dtheta]_{\times})\normal$, yielding the state:
                \begin{equation}
                    \Delta\plane = [\,\delta\dloc, \;\dtheta^\intercal\,]^\intercal \in \mathbb{R}^4 .
                    \label{eq:state}
                \end{equation}
                To avoid minimal-chart singularities or tangent-basis bookkeeping~\cite{Simultaneous-Localization-and-Mapping-with-Infinite-Planes}, the unobservable rotation about $\normal$ is constrained via the angular prior $\Sigma_{\dtheta} = \sigma_{\theta_0}^2( \mathbf{I} - \normal\normal^\intercal ) + \epsilon\,\sigma_{\theta_0}^2\, \normal\normal^\intercal$, with $\epsilon \ll 1$.
                
                \subsubsection{Continuous Occupancy Field}
                The spatial boundary of each patch is represented not as an explicit polygon, but as a continuous occupancy field over the supporting plane.
                Whereas an explicit polygon enforces a binary classification (inside versus outside), an occupancy field models partial occupancy and boundary uncertainty as a smooth, probabilistic continuum.
                Furthermore, unlike occupancy grids, which require committing to a fixed spatial resolution, compactly supported radial basis functions (CSRBFs) provide a continuous, meshless generalisation across the 2D plane that avoids discretisation artefacts.
                Writing $\uv \in \mathbb{R}^2$ for coordinates in the basis $\basis$ centred at $\pivot$, the patch carries a continuous field:
                \begin{equation}
                    \occ_k(\uv) = \logistic\Bigl( \logodds_0 + \sum_{m} \amp_m \, \kernel_m(\uv) \Bigr),
                    \quad
                    \logistic(z) = \frac{1}{1+e^{-z}},
                    \label{eq:field}
                \end{equation}
                where the sum runs over CSRBF kernels $\kernel_m$, implemented as truncated Gaussians with centres $\uv_m$, cutoff radii $\krad_m$, and amplitudes $\amp_m$.
                The sum evaluates a log-odds field: $\logodds_0$ is the prior logit of unobserved space, the logistic $\logistic$ maps it to a probability, and evidence accumulates additively in $\logodds$ in the same manner as an occupancy grid.
                The physical patch boundary is defined by the $0.5$ iso-contour of $\occ_k$ ($\logodds \geq 0$), extracted on export via marching squares.
                Because each kernel $\kernel_m$ vanishes outside its cutoff radius $\krad_m$, \refeq{eq:field} evaluates only locally overlapping kernels, enabling patches of arbitrary spatial extent to be evaluated at $\mathcal{O}(1)$ cost.

        \subsection{Ray Traversal: Sensor Model and Spatial Acceleration}\label{sec:search}
                Each LiDAR return is modelled as a directed ray $(\origin, \direction, \range)$ with origin positional and orientational uncertainties $(\sigma_{\origin}^2, \sigma_{\direction}^2)$, and range measurement noise $\sigma_{\range}^2$. Preserving line-of-sight geometry captures sensor physics directly, enabling range-space likelihoods and ray clearance to be evaluated without voxelisation. To account for sensor motion during continuous scan acquisition, each point cloud return is motion-undistorted to recover its individual ray origin $\origin$. 
                Before scoring, each ray retrieves intersecting candidate patches $\mathcal{C} \subset \patchset$. We support two spatial query backends: a CPU backend organising the $\nkernel$ active CSRBF kernels into a dynamic bounding volume hierarchy (BVH)~\cite{Ray-tracing-complex-scenes} queried at $\mathcal{O}(\log \nkernel)$ expected cost per ray, and a GPU backend projecting scan rays and CSRBF kernels onto a spherical azimuth-elevation buffer~\cite{SuMa} of angular resolution $\nazim \times \nelev$ to retrieve depth-sorted candidates at $\mathcal{O}(\nkernel + \nazim\nelev)$ cost.

        \subsection{Probabilistic Ray Assignment}\label{sec:assignment}
                For each incoming ray, the system must determine which candidate surfaces it reinforces, or whether it discovers an unmodelled structure. Rather than making hard assignments that become ambiguous at surface intersections, we formulate ray association as inference in a Dirichlet process Gaussian mixture model (DP-GMM)~\cite{The-infinite-Gaussian-mixture-model}, where candidate patches act as mixture components competing to explain the observed range.
                Inference proceeds through four sequential steps:

                \subsubsection{Range-Space Predictive Likelihood}
                    Each ray $(\origin, \direction, \range)$ intersects a single candidate patch $\plane = (\normal, \dloc, \pivot)$ at an expected range of $f(\plane, \origin, \direction) = \bigl(\dloc - \normal^\intercal (\origin - \pivot)\bigr)/(\normal^\intercal \direction)$. Linearising $f$ about the predicted hit $\hit = \origin + \expectedrange \direction$ under plane covariance $\patchcov$ and pose noise $(\sigma_{\origin}^2, \sigma_{\direction}^2)$ yields a 1D predictive Gaussian along the ray:
                    \begin{equation}
                        P(\range \mid \plane_k) \approx \mathcal{N}(\range \mid \expectedrange_k, \sigma_{\expectedrange_k}^2), \qquad \expectedrange_k = f(\plane_k, \origin, \direction).
                        \label{eq:predictive}
                    \end{equation}
                    With pose Jacobians $J_{\origin} = -\frac{\normal^\intercal}{\normal^\intercal \direction}$ and $J_{\direction} = \frac{\expectedrange (\normal \times \direction)^\intercal}{\normal^\intercal \direction}$, the plane Jacobian with respect to state, \refeq{eq:state}, is
                    \begin{equation}
                            J_{\plane} = \frac{1}{\normal^\intercal \direction} \left[\, 1, \;\; -\bigl(\normal \times (\hit - \pivot)\bigr)^\intercal \,\right],
                            \label{eq:J_plane}
                    \end{equation}
                    whose angular block is orthogonal to $\normal$. Adding surface roughness $\sigma_{\roughness}^2$ to account for imperfect planarity, the total innovation variance along the ray is
                    \begin{equation}
                            \sigma_{\expectedrange_k}^2 = \sigma_{\range}^2 + \sigma_{\roughness}^2 + J_{\plane} \patchcov[_k] J_{\plane}^\intercal + \sigma_{\origin}^2 \|J_{\origin}\|^2 + \sigma_{\direction}^2 \|J_{\direction}\|^2 .
                            \label{eq:sigma_rbar}
                    \end{equation}
                    Together, \refeq{eq:predictive} and \refeq{eq:sigma_rbar} project all 3D sensor, pose, and geometric uncertainties onto a 1D range distribution along the beam, evaluating the likelihood that candidate patch $k$ explains the return. Before computing posterior assignments, this measurement likelihood is combined with a prior reflecting each patch's geometric certainty, spatial occupancy support, and viewing geometry.

                \subsubsection{Geometric Certainty Prior}
                    In standard DP-GMM clustering, the prior scales with cluster popularity ($P(k) \propto n_k$)~\cite{The-infinite-Gaussian-mixture-model}, using point count as an implicit proxy for parameter certainty. For planar surfaces, however, count reflects certainty only in position; a single LiDAR sweep samples points along a narrow strip; fitting planes to such a strip leaves the plane's normal orientation under-constrained. Furthermore, raw counts ignore surface planarity and spatial boundaries. We therefore replace point counts with a geometric certainty prior:
                    \begin{equation}
                        P(\plane_k) \propto
                        \frac{\occ_k(\uv)^{1/4} \; \max(0, -\direction^\intercal \normal_k)}
                             {\sigma_{\distance,k} \; \sigma_{\theta,k} \; \sigma_{\mathrm{res},k}^{4}} .
                        \label{eq:prior}
                    \end{equation}
                    In the denominator, the distance and angular standard deviations ($\sigma_{\distance,k}, \sigma_{\theta,k}$) decouple positional confidence from orientation, down-weighting degenerate patches until multi-view observations constrain the full 3D plane. The residual spread $\sigma_{\mathrm{res},k}$ is the responsibility-weighted root-mean-square of the range residuals $\range - \expectedrange_k$ assigned to patch $k$. This empirical planarity metric prevent a patch from rounding over corners to form artificially ``thick'' planes.
                    In the numerator, the incidence cosine $\max(0, -\direction^\intercal \normal_k)$ accounts for viewing geometry~\cite{A-volumetric-method-for-building-complex-models-from-range-images}, suppressing grazing returns and eliminating back-facing surfaces. Finally, evaluating the continuous occupancy field $\occ_k(\uv)$ at the ray intersection prevents planes from bridging across free space; the fourth root softens boundary transitions to accommodate incremental growth.

                \subsubsection{Tempered Normalised Responsibility}
                    Combining the predictive likelihood, \refeq{eq:predictive}, with the prior, \refeq{eq:prior}, yields the posterior probability that candidate patch $k$ produced the measurement:
                    \begin{equation}
                        \resp_k = \frac{ \left( \mathcal{N}(\range \mid \expectedrange_k, \sigma_{\expectedrange_k}^2) P(\plane_k) \right)^\temp }{ \sum_j \left( \mathcal{N}(\range \mid \expectedrange_j, \sigma_{\expectedrange_j}^2) P(\plane_j) \right)^\temp } .
                        \label{eq:responsibility}
                    \end{equation}
                    We sharpen the posterior with a tempering exponent $\temp > 1$ before normalising, so that the best-fitting patch absorbs most of the responsibility. Since responsibility both scales the plane update (\refsec{sec:geometry}) and drives the occupancy field (\refsec{sec:occupancy}), this tempering exponent compounds over successive scans: the dominant patch's parameters converge, which increases its prior, \refeq{eq:prior}, and extends its boundary across the surface, while the patches it outcompetes are carved away by the exclusion term, \refeq{eq:p_out}. An entire wall is thereby consolidated into a single patch rather than fragmented into multiple patches, preserving map compactness. Crucially, $\temp$ sharpens the posterior without collapsing it: a sensor return that does not distinctly separate its candidates remains shared, ensuring assignments remain distributed when observations are ambiguous.

                \subsubsection{New Patch Creation}
                A hypothetical new patch is evaluated by assuming it is instantiated on the ray itself, explaining the measurement with $\expectedrange_{\mathrm{new}} = \range$ and $\sigma_{\expectedrange_{\mathrm{new}}}^2 = \sigma_{\range}^2 + \sigma_{\roughness}^2$. It competes in \refeq{eq:responsibility} using the uninformative prior of a newly initialised plane seeded with raw measurement noise and a broad \ang{45} angular prior. A new patch thus captures significant responsibility only when all existing patches explain the range measurement poorly. When $\resp_{\mathrm{new}}$ exceeds a threshold $\respmin$, a new patch is initialised and added to $\patchset$.

        \subsection{Incremental Update: Supporting Plane}\label{sec:geometry}
                Following ray-to-patch assignment, each ray recursively updates the associated candidate patches. Rather than accumulating measurements for offline batch fitting, we update each patch's supporting plane using an extended Kalman filter (EKF)~\cite{Probabilistic-robotics}, maintaining a Gaussian belief $\plane \sim \mathcal{N}(\plane^*, \patchcov)$ that remains valid after every scan.
                Each update is linearised at the current estimate using the plane Jacobian, \refeq{eq:J_plane}. The measurement noise corresponds to \refeq{eq:sigma_rbar} excluding the plane covariance term, which is supplied by the filter state covariance. The measurement information is scaled by $\resp_k$, following standard expectation-maximisation weighting for a mixture of components.
                This produces an incrementally refined Gaussian estimate of each plane, maintaining geometry without accumulating historical point clouds.

        \subsection{Incremental Update: Plane Boundary}\label{sec:occupancy}
                With the 3D supporting plane updated, the remaining task is to determine where the physical surface actually exists. Using the ray responsibilities $\resp_k$, we update the patch's continuous occupancy field $\occ_k$ to grow observed surfaces and carve away empty space.

                \subsubsection{Occupancy Computation}
\begin{figure}[t]
                        \centering
                        \includegraphics[width=\columnwidth]{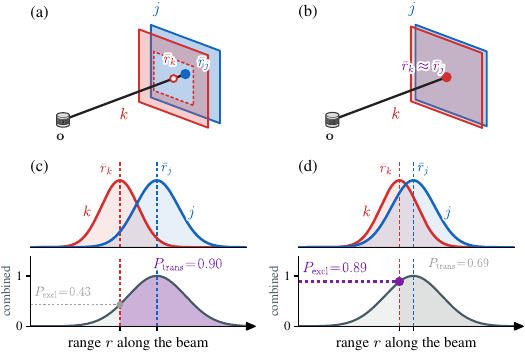}
                        \caption{Ray clearance mechanisms of \refeq{eq:p_out} across two geometric scenarios: (a, c) The beam passes through an opening in patch $k$ to land on a farther surface $j$. The separated range Gaussians yield high \textit{transmission}, carving free space on $k$; (b, d) Patches $k$ and $j$ are coplanar patches at the same depth. The overlapping range Gaussians yield high \textit{exclusion}, suppressing duplicate planes.}
                        \label{fig:clearance}
                \end{figure}

                Patch boundaries are updated using rays terminating on the patch ($\pin{k} = \resp_k$) or passing through it ($\pout{k}$). The out-probability accumulates evidence from competing candidates $j \neq k$ via transmission $\ptrans$ (free space in front of $j$) and mutual exclusion $\pexcl$ (overlapping patches):
                \begin{equation}
                    \pout{k} = \sum_{j \neq k} \resp_j \max \left( \ptrans(k, j), \, \pexcl(k, j) \right),
                    \label{eq:p_out}
                \end{equation}
                where $\ptrans(k, j) = \cdf\bigl( (\expectedrange_j - \expectedrange_k)/\sqrt{\sigma_{\expectedrange_k}^2 + \sigma_{\expectedrange_j}^2} \bigr)$, with $\cdf$ denoting the standard normal cumulative distribution function, and $\pexcl(k, j) = \exp\bigl(-\frac{(\expectedrange_j - \expectedrange_k)^2}{2(\sigma_{\expectedrange_k}^2 + \sigma_{\expectedrange_j}^2)}\bigr)$, the normalised overlap $\int \mathcal{N}(\range \mid \expectedrange_k, \sigma_{\expectedrange_k}^2)\,\mathcal{N}(\range \mid \expectedrange_j, \sigma_{\expectedrange_j}^2)\,\mathrm{d}\range$ of the two predictive range distributions (\reffig{fig:clearance}). Both terms share the combined range variance $\sigma_{\expectedrange_k}^2 + \sigma_{\expectedrange_j}^2$, so neither requires an additional scaling parameter.

                We form the continuous hit/miss observation pair $(w_k, y_k)$ at predicted hit coordinates $\uv$:
                \begin{equation}
                    w_k = \pin{k} + \pout{k},
                    \qquad
                    y_k = \frac{\pin{k}}{w_k},
                    \label{eq:bayes_pair}
                \end{equation}
                where total confidence $w_k \in [0, 1]$ weights filter updates and target $y_k \in [0, 1]$ generalises occupancy grid updates~\cite{OctoMap} to continuous hit/miss evidence. When $w_k \geq 0.5$, this observation pair is applied to update the field, mapping the assignment $\resp_k$ to target occupancy with $y_k \approx 1$ for solid surface and $y_k \approx 0$ for carved free space.

                \subsubsection{Online Boundary Learning via Modified GGAP-RBF}
                The occupancy field, \refeq{eq:field}, is updated online using GGAP-RBF~\cite{GGAP-RBF}, which dynamically allocates and refines local kernels via recursive EKF updates.
                We modify both allocation and gating to depend on the sample's in-plane position uncertainty $\delta$, computed by projecting the sensor pose and beam pointing uncertainties along the ray onto the supporting plane:
                \begin{equation}
                    \delta = \frac{\sqrt{\sigma_{\origin}^2 + \expectedrange^2\sigma_{\direction}^2}}{|\direction^\intercal\normal|} .
                    \label{eq:in_plane_drift}
                \end{equation}

                \emph{a) Adaptive Allocation:} we replace the fixed centre-distance floor $\epsilon_n$ with $d^* > \max(\epsilon_n, \delta)$, preventing distant or grazing returns from forming premature edges before sufficient local evidence is gathered.

                \emph{b) Pose-Drift Gating:} to prevent pose jitter and beam divergence from dilating boundaries outward across silhouettes, confidence $w_k$ is gated by $\delta$: it retains full weight for $\delta \leq \delta_{\mathrm{ref}}$, drops to zero for $\delta > \delta_{\mathrm{cut}}$, and decays quadratically as $(\delta_{\mathrm{ref}}/\delta)^{2}$ between the two thresholds.

                Together, these mechanisms make boundary learning uncertainty-aware: fine-grained kernels and crisp edges are asserted only when the sensor observes a surface closely and perpendicularly ($\delta$ is small), while distant or grazing observations are prevented from inflating boundaries into empty space.

        \subsection{Texture Projection}\label{sec:texture}
                The final stage of the pipeline projects visual appearance onto the reconstructed geometry. Colour pixels from RGB images $\image_t$ are back-projected onto 2D texture atlases $\atlas_k$ for each 3D patch. Rather than standard Z-buffering, which causes colour bleeding across noisy or overlapping boundaries, candidate surfaces along each camera ray are sorted in an A-buffer~\cite{The-A-buffer-an-antialiased-hidden-surface-method}. The visibility of patch $k$ is composed as a transmittance product $\pvis = \occ_k \prod_{j < k} (1 - \occ_j)$ over preceding surfaces $j$. Observations are accumulated into each patch atlas weighted by visibility $\pvis$, the projected texel footprint, and the inverse covariance of sensor pose and plane geometry. \reffig{fig:banner_long} illustrates textured reconstructions qualitatively. A quantitative evaluation of appearance is deferred to future work.

\section{EXPERIMENTS}
        \subsection{Baselines and Experimental Setup}\label{sec:setup}

                We compare OPP against four distinct LiDAR reconstruction baselines, each evaluated at a \SI{10}{\centi\metre} resolution: Poisson meshing (Puma~\cite{Puma}), neural signed distance fields (PINGS~\cite{PINGS}), direct voxelised meshing (ImMesh~\cite{ImMesh}), and truncated signed distance fields (VDBFusion~\cite{VDBFusion}). All pipelines use identical ground-truth trajectories to isolate reconstruction fidelity from odometry drift, with ImMesh adapted to accept external poses. We omit PlanarMesh~\cite{PlanarMesh} because it was designed as an in-memory proof-of-concept without paging and exhausts system memory after approximately 300 scans; in contrast, OPP uses disk-backed paging (\reffig{fig:memory_time}) to maintain a strictly bounded memory footprint.

                Geometric fidelity is evaluated against millimetre-accurate reference scans from a survey-grade tripod scanner (Leica RTC360). Runtime and memory benchmarks (\reffig{fig:memory_time}, \reffig{fig:frame_time_hist}) were recorded on an Apple M2 laptop (8-core CPU, 10-core GPU, \SI{24}{\giga\byte} unified memory).
                                
                \begin{table*}[ht]
                        \centering
                        \caption{Quantitative performance across benchmark sequences, evaluated using the metrics defined in \refsec{sec:datasets_metrics}. \textit{Accuracy} and \textit{completeness} report mean and mode distance errors against ground truth; precision, recall, and F1 are evaluated at \SI{1}{\centi\metre} (strict surface localisation) and \SI{5}{\centi\metre}. Bold indicates the best and underline indicates the second best. We reiterate that our approach is optimised for \textit{accuracy} rather than unconstrained \textit{completeness}.}
                        \label{tab:quantitative_results}
                        \resizebox{\textwidth}{!}{
                        \begin{tabular}{l | l | c c c c | c c c | c c c | c}
                                \hline
                                & & & & & & \multicolumn{3}{c|}{\textbf{at \SI{1}{\centi\metre}}} & \multicolumn{3}{c|}{\textbf{at \SI{5}{\centi\metre}}} & \\
                                \textbf{Seq.} & \textbf{Method} & \textbf{Mean Acc.} & \textbf{Mean Comp.} & \textbf{Mode Acc.} & \textbf{Mode Comp.} & \textbf{Prec.} & \textbf{Rec.} & \textbf{F1} & \textbf{Prec.} & \textbf{Rec.} & \textbf{F1} & \textbf{Map Size} \\
                                & & (m) $\downarrow$ & (m) $\downarrow$ & (m) $\downarrow$ & (m) $\downarrow$ & $\uparrow$ & $\uparrow$ & $\uparrow$ & $\uparrow$ & $\uparrow$ & $\uparrow$ & (MiB) $\downarrow$ \\
                                \hline
                                \multirow{5}{*}{\shortstack[l]{Keble03\\(4.8 min)}} & Puma & \underline{0.049} & 0.824 & \textbf{0.010} & 0.011 & 12.3\% & \textbf{22.5\%} & 15.9\% & 69.6\% & 75.9\% & 72.6\% & 1,744.7 \\
                                 & PINGS & 0.092 & 0.064 & \underline{0.011} & \underline{0.009} & 7.8\% & 12.7\% & 9.7\% & 53.3\% & 63.4\% & 57.9\% & 161.0 \\
                                 & ImMesh & 0.061 & \textbf{0.033} & 0.012 & \textbf{0.008} & 10.9\% & \underline{21.3\%} & 14.5\% & 69.8\% & \textbf{85.6\%} & \underline{76.9\%} & 154.3 \\
                                 & VDBFusion & 0.052 & \underline{0.043} & \underline{0.011} & \underline{0.009} & \underline{15.4\%} & 20.2\% & \underline{17.4\%} & \underline{76.7\%} & \underline{79.8\%} & \textbf{78.2\%} & \underline{30.6} \\
                                 & \textbf{OPP (Ours)} & \textbf{0.030} & 1.181 & \textbf{0.010} & \textbf{0.008} & \textbf{16.8\%} & 18.9\% & \textbf{17.7\%} & \textbf{86.2\%} & 62.2\% & 72.3\% & \textbf{30.2} \\
                                \hline
                                \multirow{5}{*}{\shortstack[l]{Observatory01\\(4.8 min)}} & Puma & 0.086 & 0.125 & \textbf{0.007} & \textbf{0.006} & 10.3\% & \textbf{48.5\%} & \underline{17.0\%} & 50.6\% & \textbf{81.1\%} & 62.3\% & 1,698.5 \\
                                 & PINGS & 0.152 & 0.070 & \underline{0.010} & 0.009 & 5.5\% & 10.0\% & 7.1\% & 33.9\% & 54.5\% & 41.8\% & 265.6 \\
                                 & ImMesh & 0.103 & \textbf{0.038} & 0.013 & \underline{0.008} & 8.0\% & 19.9\% & 11.4\% & 48.8\% & \underline{79.9\%} & 60.6\% & 242.6 \\
                                 & VDBFusion & \underline{0.085} & \underline{0.054} & \textbf{0.007} & 0.009 & \underline{12.8\%} & 15.4\% & 14.0\% & \underline{58.3\%} & 68.2\% & \underline{62.9\%} & \textbf{50.0} \\
                                 & \textbf{OPP (Ours)} & \textbf{0.039} & 0.506 & \textbf{0.007} & \underline{0.008} & \textbf{19.1\%} & \underline{20.4\%} & \textbf{19.7\%} & \textbf{79.2\%} & 66.3\% & \textbf{72.2\%} & \underline{50.5} \\
                                \hline
                                \multirow{5}{*}{\shortstack[l]{Bodleian02\\(8.3 min)}} & Puma & 0.098 & 0.578 & \textbf{0.010} & 0.012 & 8.9\% & 12.2\% & 10.3\% & 57.8\% & 62.6\% & 60.1\% & 1,244.7 \\
                                 & PINGS & 0.114 & 0.203 & \underline{0.011} & \underline{0.009} & 8.6\% & 13.7\% & 10.5\% & 54.0\% & 62.0\% & 57.8\% & 409.5 \\
                                 & ImMesh & 0.100 & \underline{0.145} & \textbf{0.010} & \textbf{0.008} & 10.4\% & \textbf{16.9\%} & \underline{12.9\%} & 60.6\% & \underline{67.6\%} & 63.9\% & 285.2 \\
                                 & VDBFusion & \underline{0.095} & \textbf{0.056} & \underline{0.011} & \underline{0.009} & \underline{10.8\%} & 16.0\% & \underline{12.9\%} & \underline{64.4\%} & \textbf{70.9\%} & \textbf{67.5\%} & \underline{113.7} \\
                                 & \textbf{OPP (Ours)} & \textbf{0.075} & 1.461 & \textbf{0.010} & \textbf{0.008} & \textbf{12.5\%} & \underline{16.3\%} & \textbf{14.2\%} & \textbf{75.3\%} & 57.0\% & \underline{64.8\%} & \textbf{104.6} \\
                                \hline
                                \multirow{5}{*}{\shortstack[l]{kth\_day\_10\\(10.3 min)}} & Puma & \underline{0.099} & 0.412 & 0.023 & \textbf{0.011} & 3.2\% & \textbf{20.3\%} & \textbf{5.6\%} & \underline{55.8\%} & \textbf{66.0\%} & \textbf{60.5\%} & 9,228.5 \\
                                 & PINGS & 0.227 & \underline{0.195} & 0.026 & 0.017 & 1.0\% & 2.8\% & 1.4\% & 24.1\% & 37.5\% & 29.4\% & \textbf{127.0} \\
                                 & ImMesh & 0.123 & 0.213 & \underline{0.022} & \underline{0.015} & \underline{3.4\%} & \underline{7.1\%} & \underline{4.6\%} & 54.6\% & \underline{60.5\%} & \underline{57.4\%} & 219.3 \\
                                 & VDBFusion & 0.135 & \textbf{0.082} & \textbf{0.021} & 0.016 & 2.9\% & 6.2\% & 3.9\% & 45.6\% & 58.7\% & 51.3\% & 336.1 \\
                                 & \textbf{OPP (Ours)} & \textbf{0.090} & 5.148 & \textbf{0.021} & \underline{0.015} & \textbf{3.7\%} & 3.2\% & 3.4\% & \textbf{61.3\%} & 28.2\% & 38.7\% & \underline{191.0} \\
                                \hline
                                \multirow{5}{*}{\shortstack[l]{tuhh\_day\_02\\(8.3 min)}} & Puma & \underline{0.122} & 8.250 & \textbf{0.017} & \textbf{0.012} & \textbf{4.7\%} & \textbf{13.1\%} & \textbf{6.9\%} & \underline{48.8\%} & 50.8\% & \underline{49.8\%} & 7,877.1 \\
                                 & PINGS & 0.400 & \underline{0.100} & 0.024 & 0.017 & 1.0\% & 4.1\% & 1.5\% & 18.2\% & 52.5\% & 27.0\% & \textbf{84.3} \\
                                 & ImMesh & 0.203 & 0.119 & \underline{0.018} & 0.015 & 3.0\% & \underline{9.5\%} & 4.6\% & 34.5\% & \underline{71.9\%} & 46.6\% & \underline{152.2} \\
                                 & VDBFusion & 0.225 & \textbf{0.063} & \underline{0.018} & 0.016 & 3.0\% & 8.3\% & 4.4\% & 34.9\% & \textbf{72.5\%} & 47.1\% & 221.5 \\
                                 & \textbf{OPP (Ours)} & \textbf{0.108} & 3.009 & 0.019 & \underline{0.014} & \underline{4.3\%} & 6.5\% & \underline{5.1\%} & \textbf{55.9\%} & 48.0\% & \textbf{51.7\%} & 157.5 \\
                                \hline
                                \multirow{5}{*}{\shortstack[l]{ntu\_day\_02\\(3.8 min)}} & Puma & 0.112 & 0.318 & 0.026 & \textbf{0.011} & 1.7\% & \textbf{23.2\%} & 3.2\% & 33.6\% & \textbf{74.7\%} & \underline{46.4\%} & 5,645.0 \\
                                 & PINGS & 0.239 & \underline{0.296} & 0.049 & 0.017 & 0.4\% & 3.0\% & 0.8\% & 13.0\% & 41.0\% & 19.7\% & \textbf{70.0} \\
                                 & ImMesh & \underline{0.085} & 0.321 & \textbf{0.023} & \underline{0.014} & \underline{2.8\%} & \underline{8.9\%} & \textbf{4.2\%} & \underline{41.2\%} & \underline{59.6\%} & \textbf{48.7\%} & \underline{87.4} \\
                                 & VDBFusion & 0.139 & \textbf{0.085} & 0.032 & 0.017 & 1.0\% & 4.6\% & 1.7\% & 25.2\% & 57.7\% & 35.1\% & 193.6 \\
                                 & \textbf{OPP (Ours)} & \textbf{0.074} & 4.257 & \underline{0.025} & \underline{0.014} & \textbf{3.2\%} & 4.9\% & \underline{3.9\%} & \textbf{54.5\%} & 37.1\% & 44.2\% & 98.2 \\
                                \hline
                        \end{tabular}
                        } %
                \end{table*}

                \begin{table}[t]
                        \centering
                        \caption{Parameters used during benchmarking: (a) DP-GMM (\refsec{sec:assignment}), (b) EKF (\refsec{sec:geometry}), and (c) GGAP-RBF (\refsec{sec:occupancy}).}
                        \label{tab:parameters}
                        \footnotesize
                        \setlength{\tabcolsep}{4pt}
                        \begin{tabular}{c l l l}
                                \hline
                                & \textbf{Symbol} & \textbf{Meaning} & \textbf{Value} \\
                                \hline
                                \multirow{2}{*}{(a)} & $\temp$ & tempering exponent & 2.0 \\
                                & $\respmin$ & new-patch responsibility & 0.9 \\
                                \hline
                                \multirow{4}{*}{(b)} & $\sigma_{\range}$ & sensor range noise & $1$ or $3$\,\si{\centi\metre}$^*$ \\
                                & $\sigma_{\roughness}$ & plane roughness & \SI{1}{\centi\metre} \\
                                & $\sigma_{\origin}$ & pose position uncertainty & \SI{2}{\milli\metre} \\
                                & $\sigma_{\direction}$ & pose orientation uncertainty & \SI{0.001}{\radian} \\
                                \hline
                                \multirow{2}{*}{(c)} & $\logodds_0$ & prior logit & $-5.0$ \\
                                & $\delta_{\mathrm{ref}}, \delta_{\mathrm{cut}}$ & drift gate & $2, 4$\,\si{\centi\metre} \\
                                \hline
                                \multicolumn{4}{l}{\scriptsize $^*$\SI{1}{\centi\metre} for Hesai QT64 in OSD, \SI{3}{\centi\metre} for Ouster OS1 in MCD.}
                        \end{tabular}
                \end{table}

        \subsection{Datasets and Evaluation Metrics}\label{sec:datasets_metrics}
                Experiments are conducted on six sequences from two datasets: Keble03, Observatory01, and Bodleian02 from the Oxford Spires Dataset (OSD)~\cite{Oxford-Spires-Dataset}, alongside kth\_day\_10, tuhh\_day\_02, and ntu\_day\_02 from the Multi-Campus Dataset (MCD)~\cite{Multi-Campus-Dataset}. The OSD platform provides a Hesai QT64 LiDAR and three synchronised cameras. MCD features one or two forward cameras paired with an Ouster OS1-64 LiDAR (KTH, TUHH) or an OS1-128 LiDAR (NTU).

                We measure reconstruction quality against ground-truth scans using standard distance metrics in metres: \emph{accuracy} (reconstruction-to-reference error) and \emph{completeness} (reference-to-reconstruction error), reporting both mean and mode. In addition, surface precision, recall, and F1 scores are evaluated at \SI{1}{\centi\metre} and \SI{5}{\centi\metre} thresholds. Storage requirements are assessed using each method's native disk format: meshes for Puma and ImMesh, neural weights for PINGS, volumetric grids for VDBFusion, and our compact patch format for OPP. All evaluations use a fixed configuration (\reftab{tab:parameters}) without per-sequence tuning, adjusting only range noise $\sigma_{\range}$ to match sensor specifications.

        \subsection{Reconstruction Results}\label{sec:results}
\begin{figure*}[t]
                \centering
                \newlength{\bannerw}\setlength{\bannerw}{0.180\textwidth}
                \newlength{\bannerh}\setlength{\bannerh}{0.600\bannerw}
                \newcommand{\bpanel}[1]{\parbox[c][\bannerh][c]{\bannerw}{\centering #1}}
                \newcommand{\brow}[2]{\parbox[c][\bannerh][c]{7mm}{\centering
                        \rotatebox[origin=c]{90}{\scriptsize\shortstack{#1\\#2}}}}
                \newcommand{\bhead}[1]{\parbox[b]{\bannerw}{\centering\footnotesize #1}}
                \setlength{\tabcolsep}{1.5pt}
                \renewcommand{\arraystretch}{1.1}
                \begin{tabular}{@{}m{7mm}|m{\bannerw}|m{\bannerw}|m{\bannerw}|m{\bannerw}|m{\bannerw}@{}}
                        \multicolumn{1}{c|}{} & \bhead{(a) Tripod Scan\\(Not Used)} & \bhead{(b) Raw LiDAR\\Scan}
                        & \bhead{(c) Volumetric Mesh\\(VDBFusion)} & \bhead{(d) OPP Model\\(Ours)}
                        & \bhead{(e) Textured OPP\\Model (Ours)} \\[2pt] \hline
                        \centering\brow{OSD}{Keble03}
                                & \bpanel{\includegraphics[width=\bannerw,height=\bannerh,keepaspectratio,trim=90 107 102 82,clip]{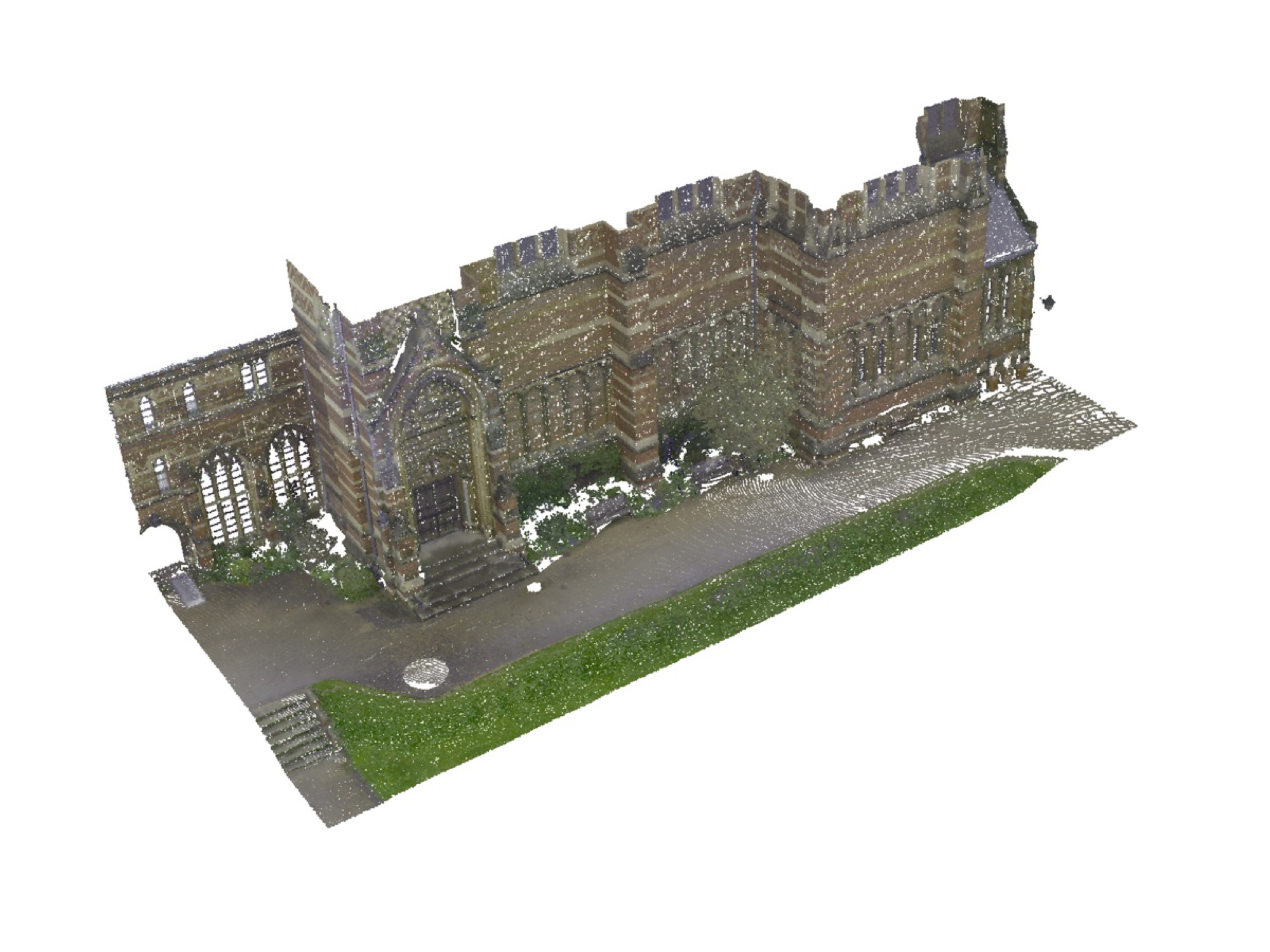}}
                                & \bpanel{\includegraphics[width=\bannerw,height=\bannerh,keepaspectratio,trim=90 107 102 82,clip]{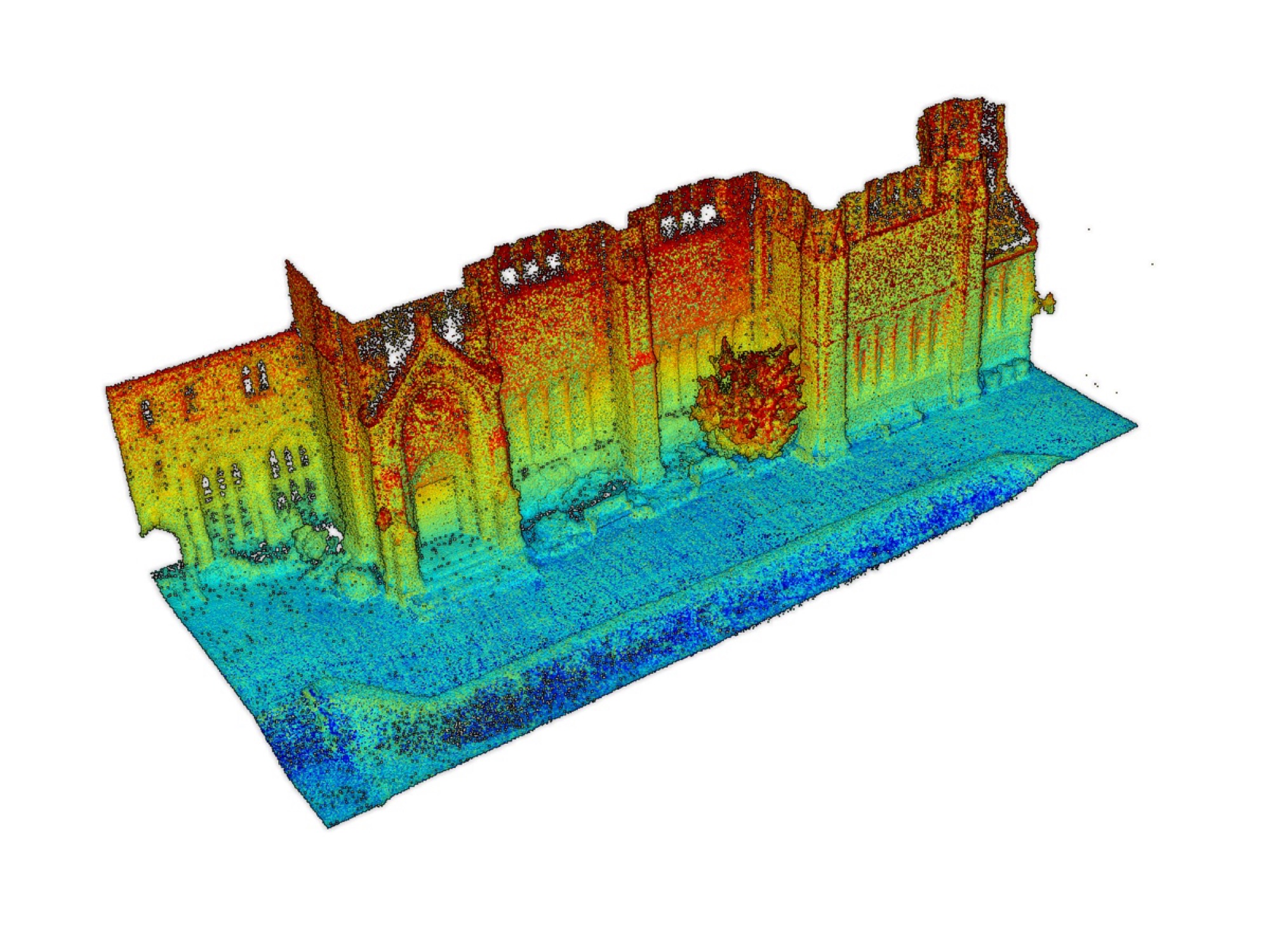}}
                                & \bpanel{\includegraphics[width=\bannerw,height=\bannerh,keepaspectratio,trim=90 107 102 82,clip]{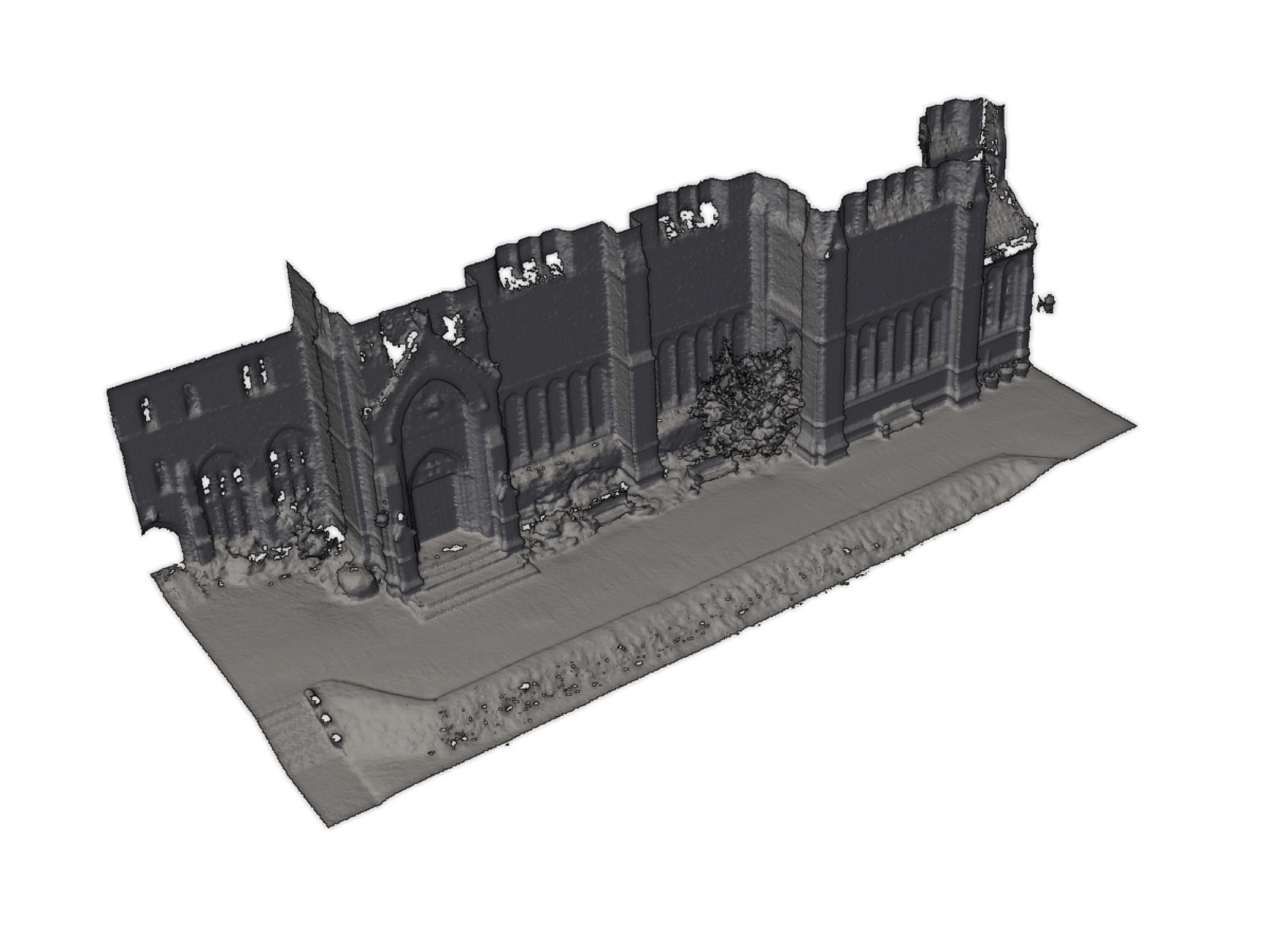}}
                                & \bpanel{\includegraphics[width=\bannerw,height=\bannerh,keepaspectratio,trim=90 107 102 82,clip]{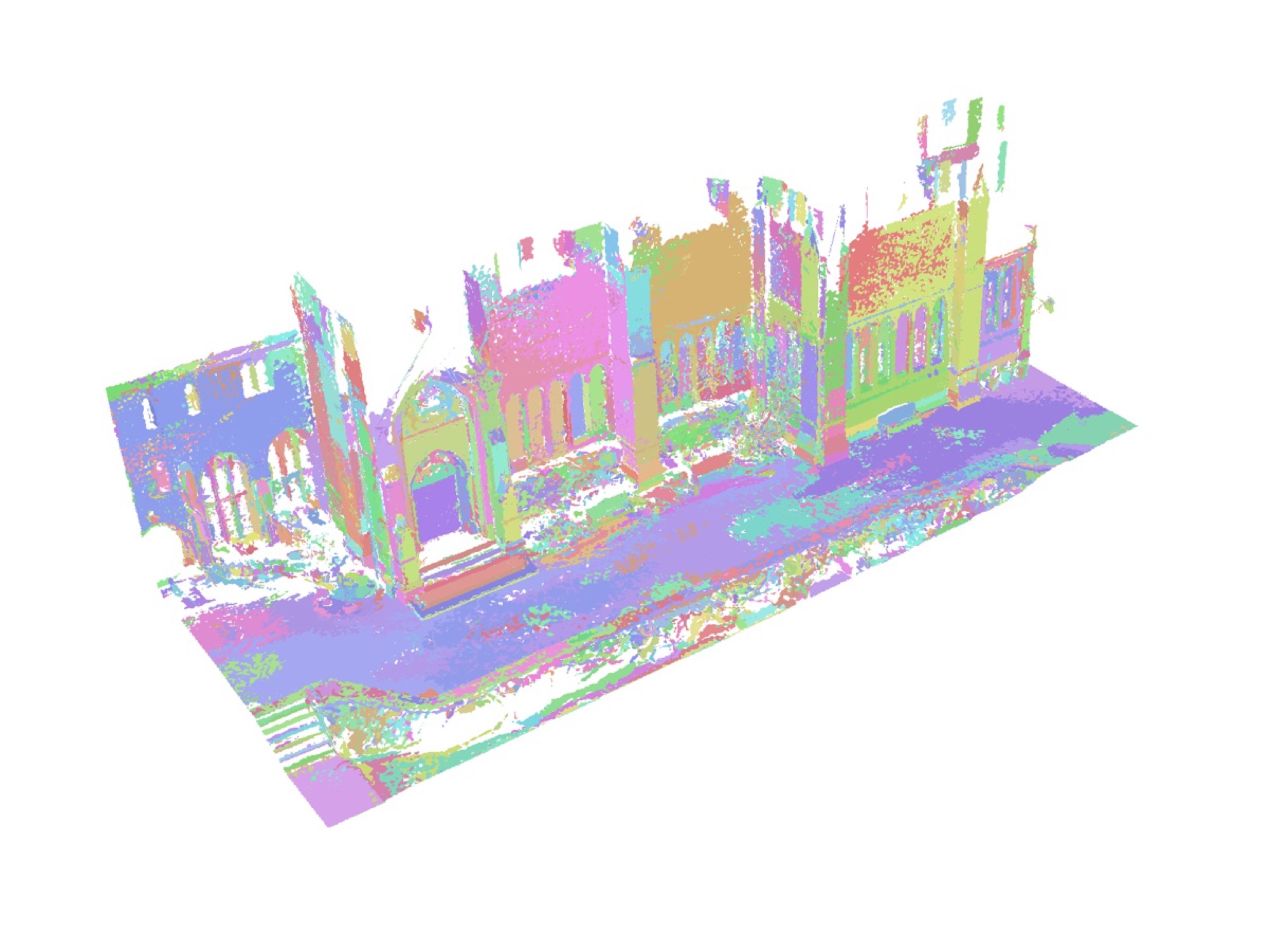}}
                                & \bpanel{\includegraphics[width=\bannerw,height=\bannerh,keepaspectratio,trim=90 107 102 82,clip]{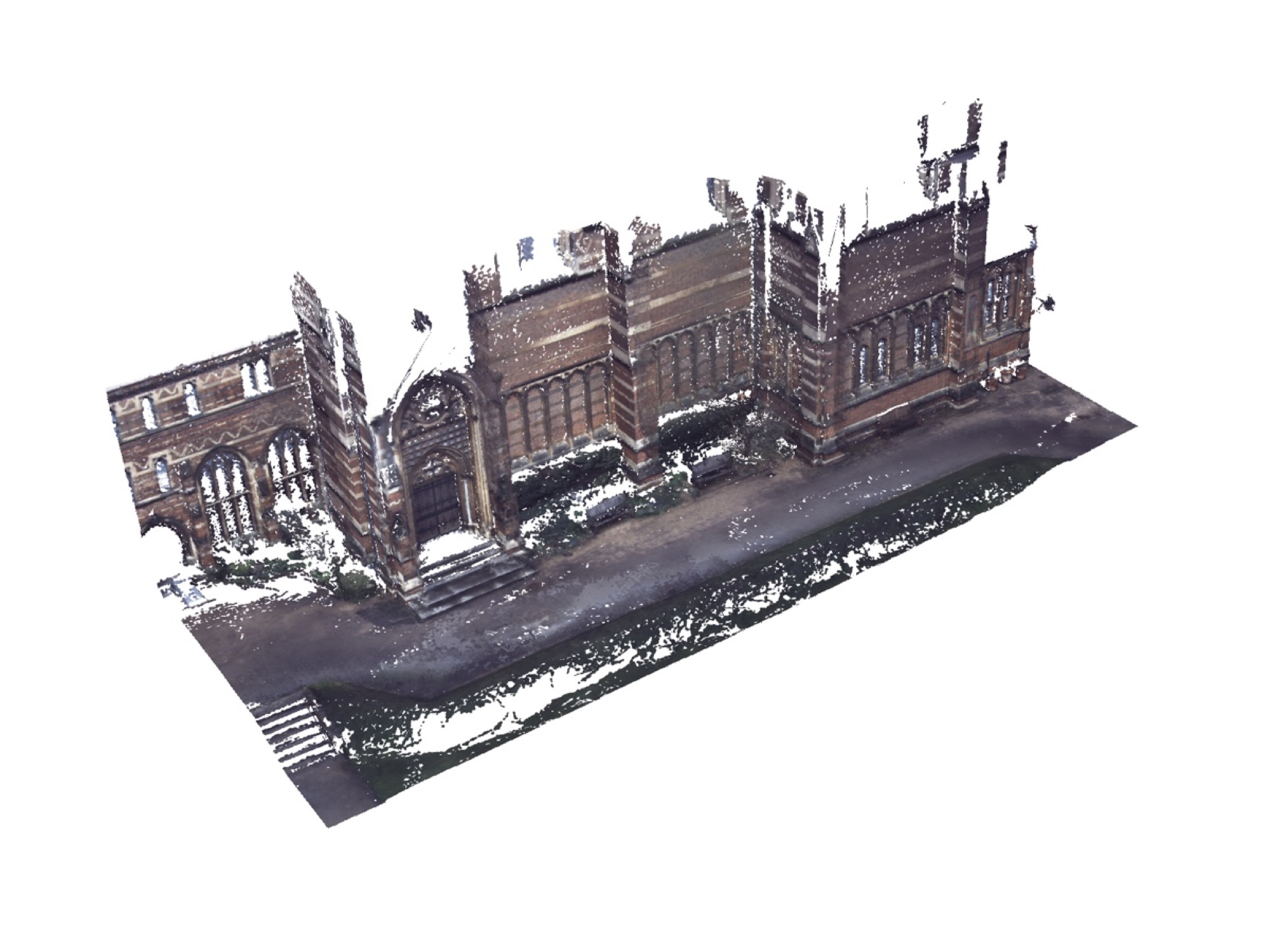}} \\ \hline
                        \centering\brow{OSD}{Bodleian02}
                                & \bpanel{\includegraphics[width=\bannerw,height=\bannerh,keepaspectratio,trim=30 147 24 64,clip]{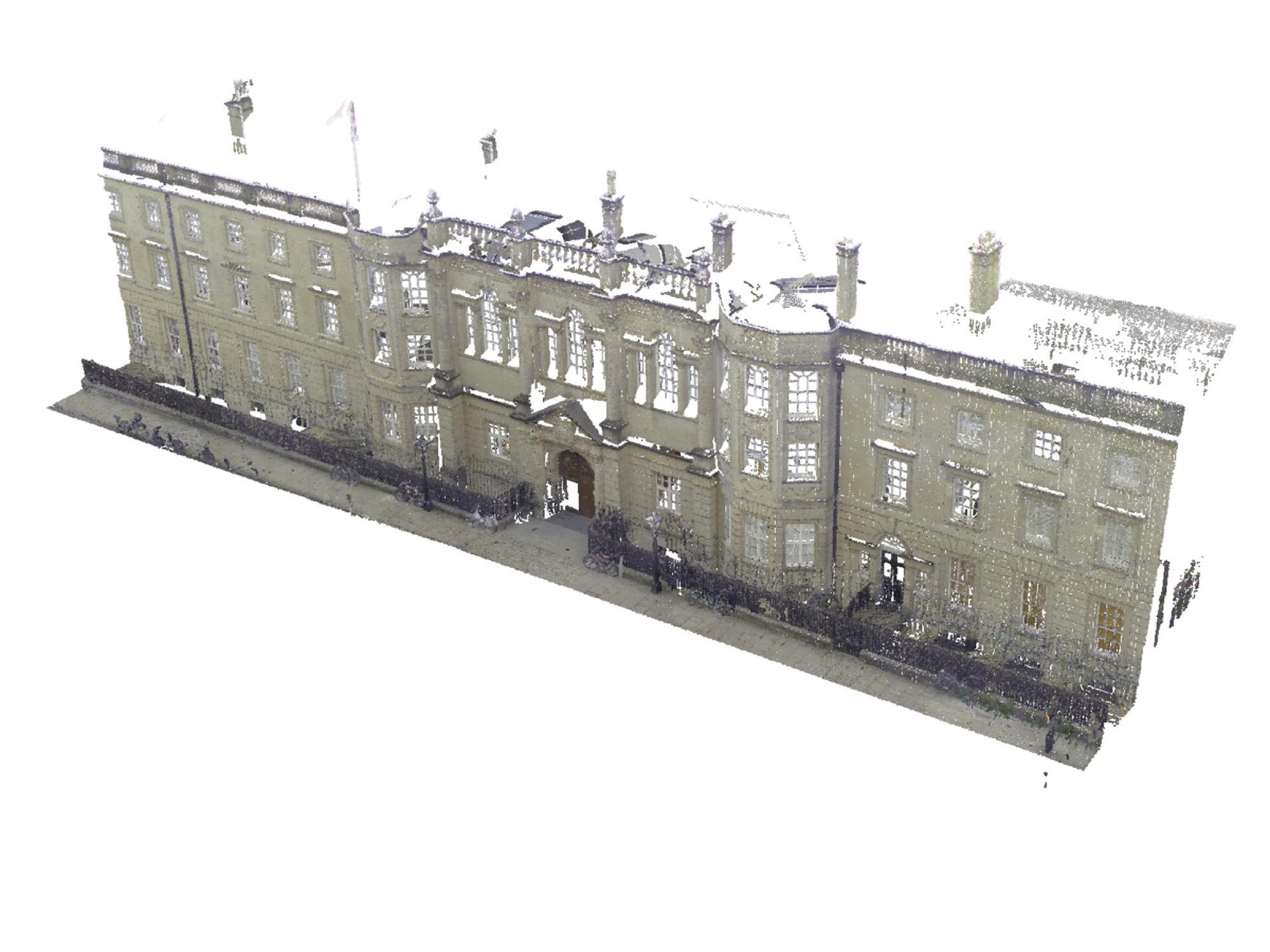}}
                                & \bpanel{\includegraphics[width=\bannerw,height=\bannerh,keepaspectratio,trim=30 147 24 64,clip]{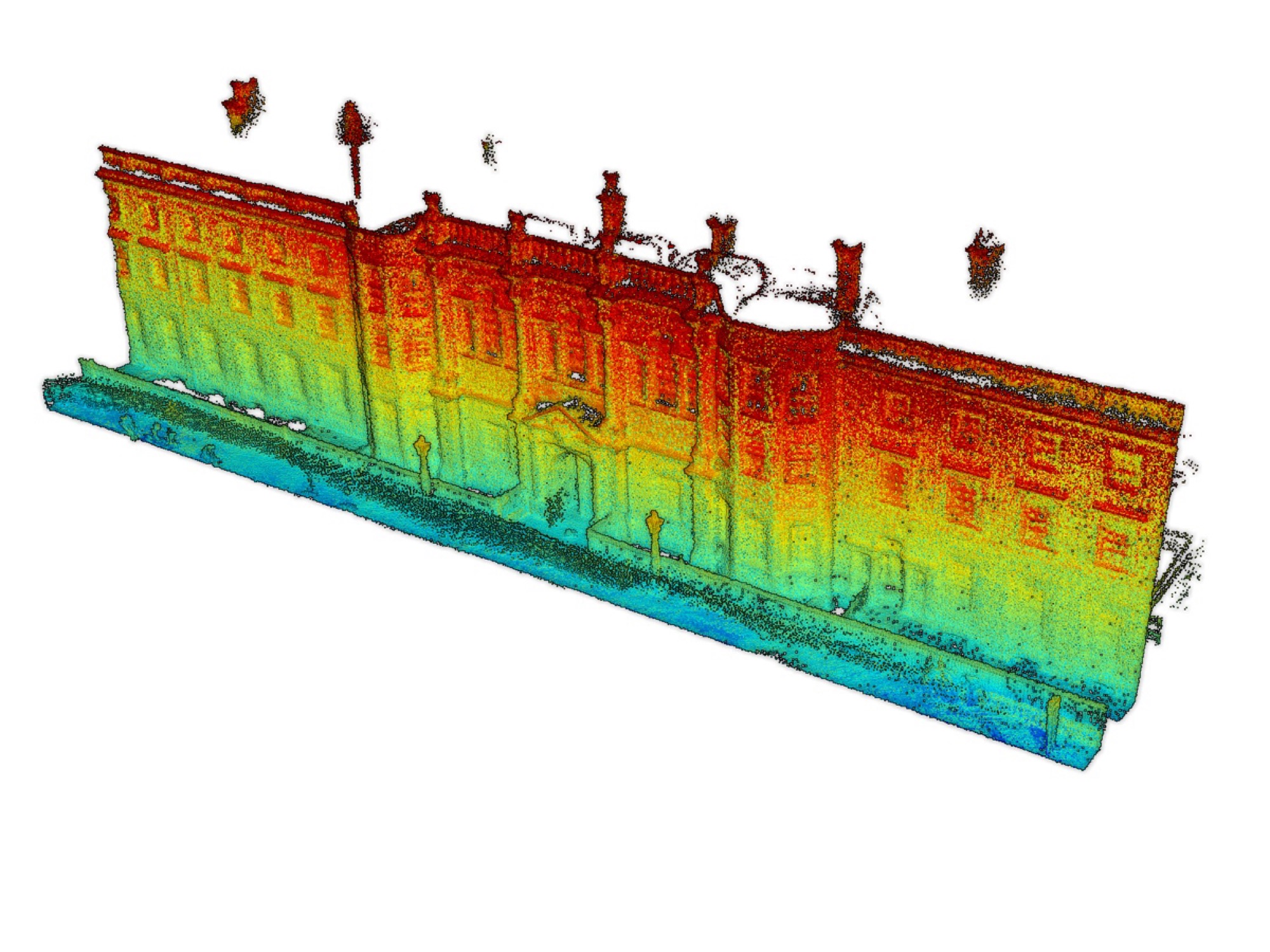}}
                                & \bpanel{\includegraphics[width=\bannerw,height=\bannerh,keepaspectratio,trim=30 147 24 64,clip]{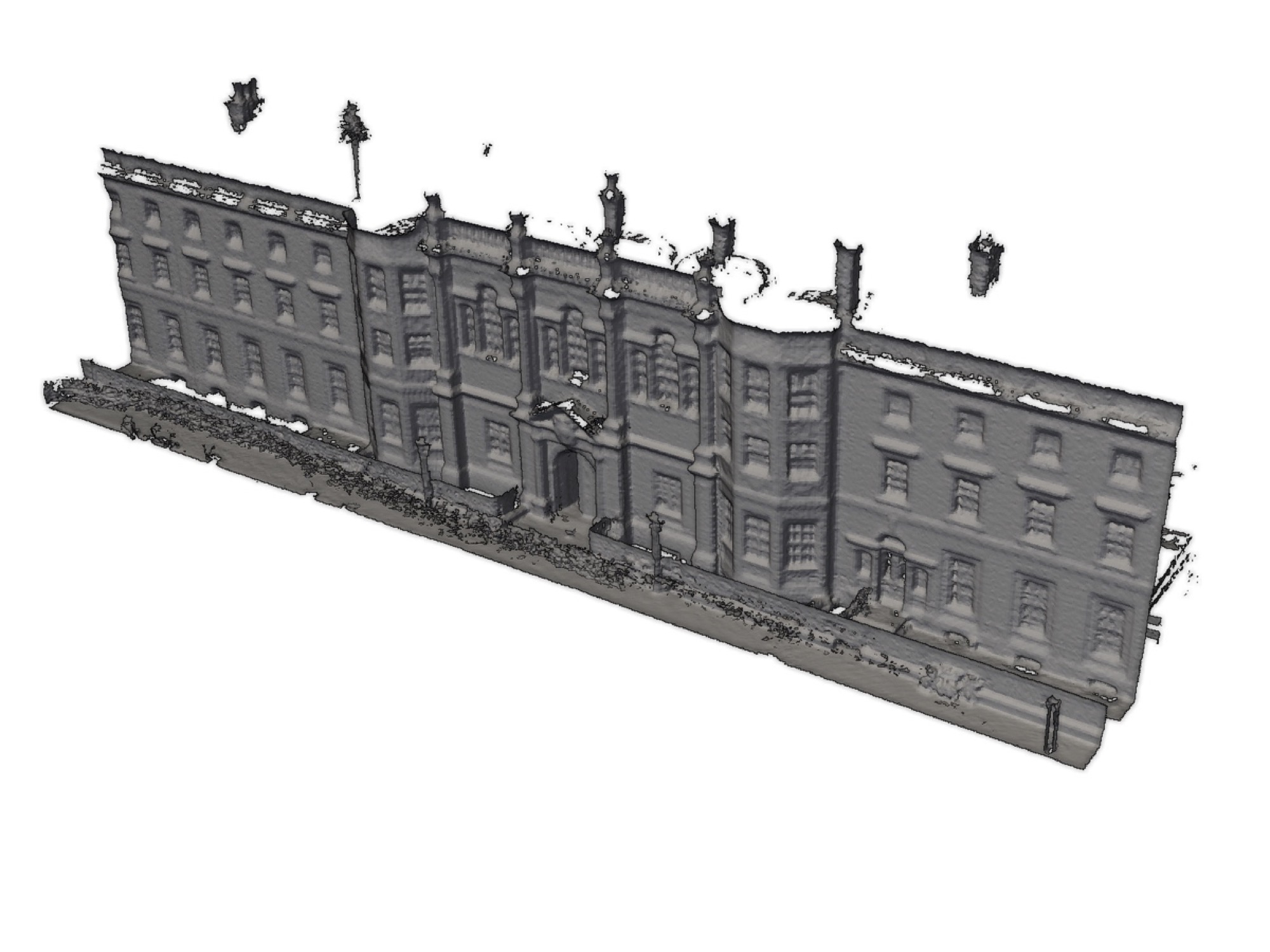}}
                                & \bpanel{\includegraphics[width=\bannerw,height=\bannerh,keepaspectratio,trim=30 147 24 64,clip]{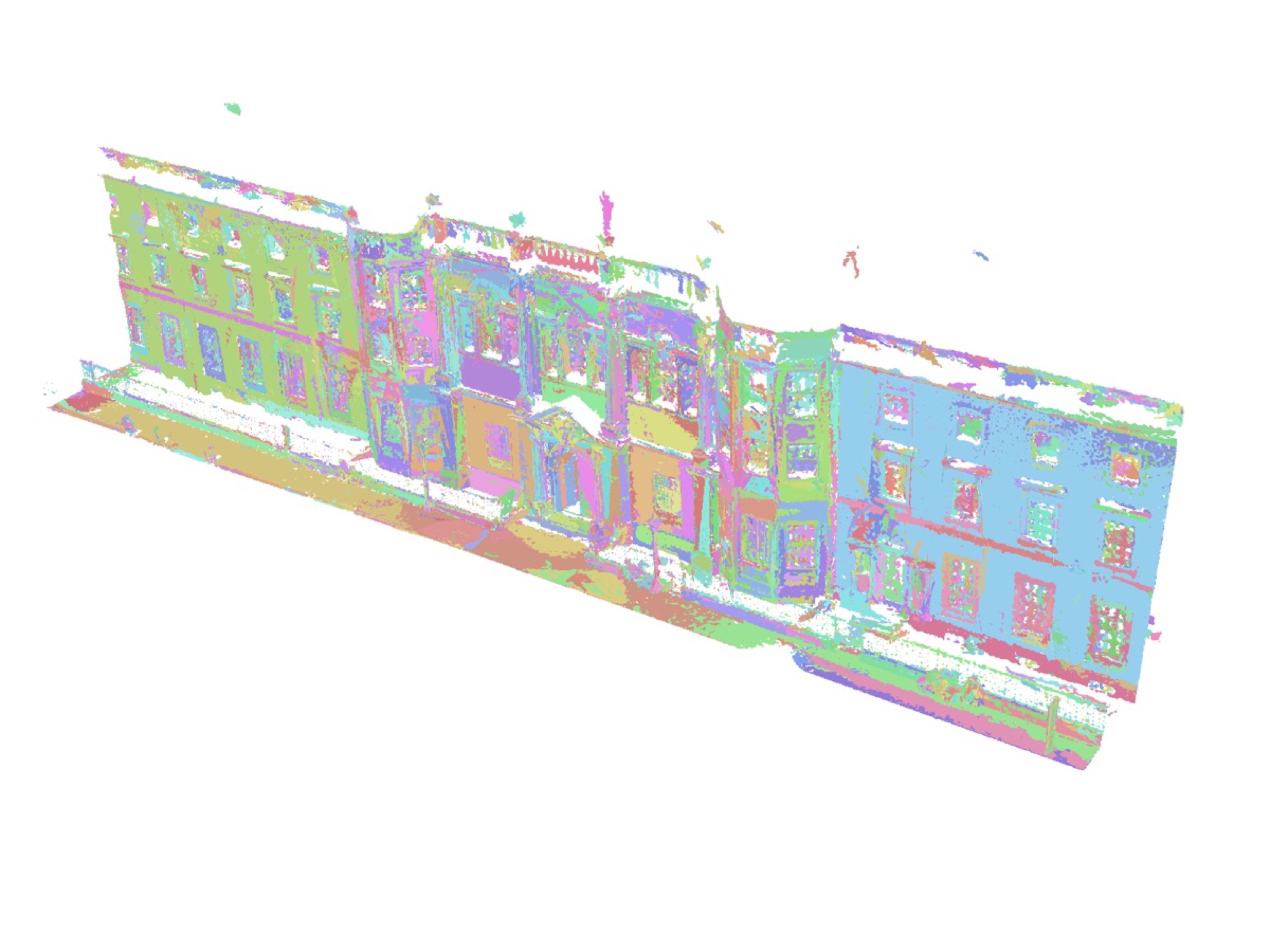}}
                                & \bpanel{\includegraphics[width=\bannerw,height=\bannerh,keepaspectratio,trim=30 147 24 64,clip]{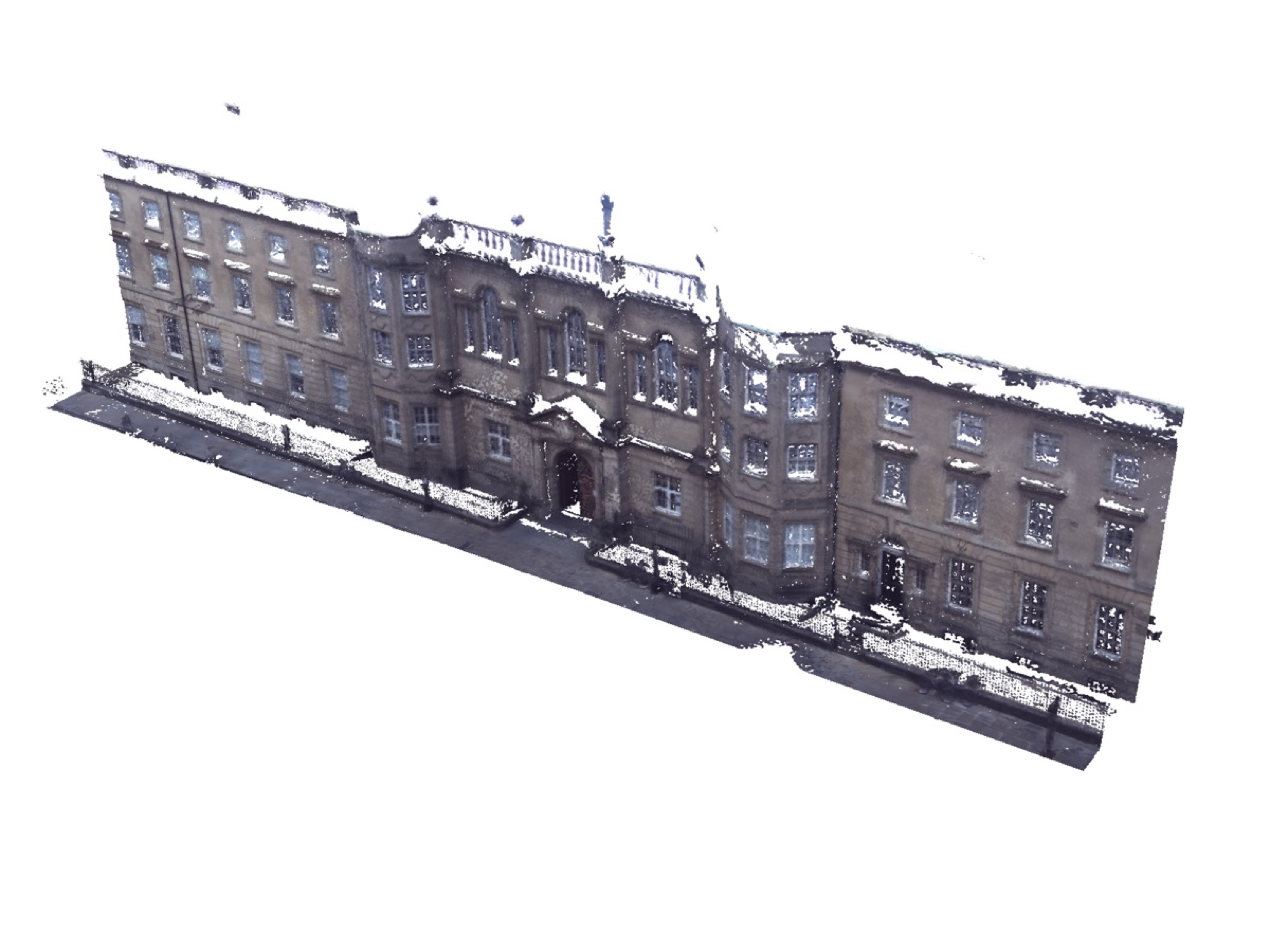}} \\ \hline
                        \centering\brow{MCD}{kth\_day\_10}
                                & \bpanel{\includegraphics[width=\bannerw,height=\bannerh,keepaspectratio,trim=84 20 41 12,clip]{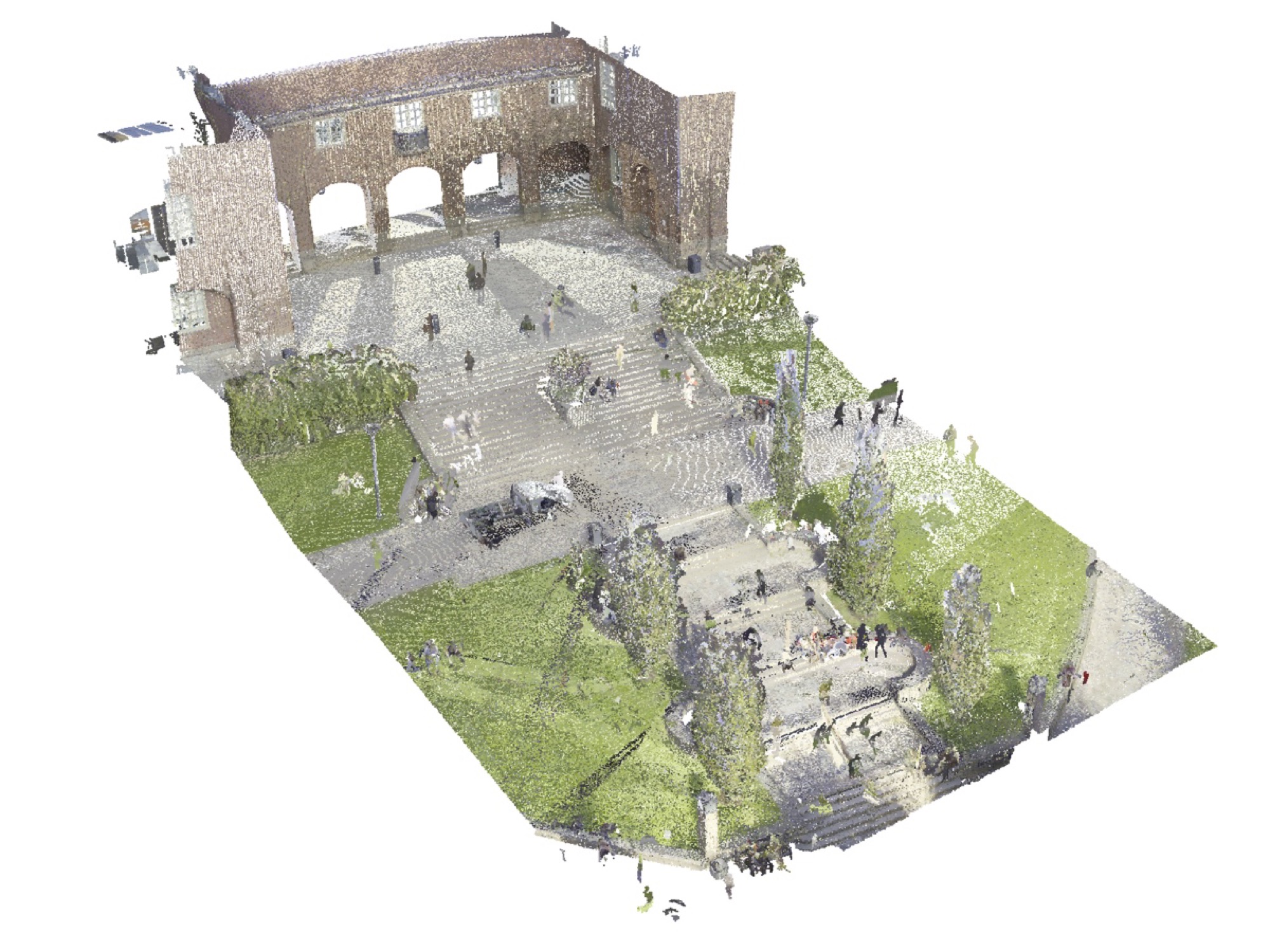}}
                                & \bpanel{\includegraphics[width=\bannerw,height=\bannerh,keepaspectratio,trim=84 20 41 12,clip]{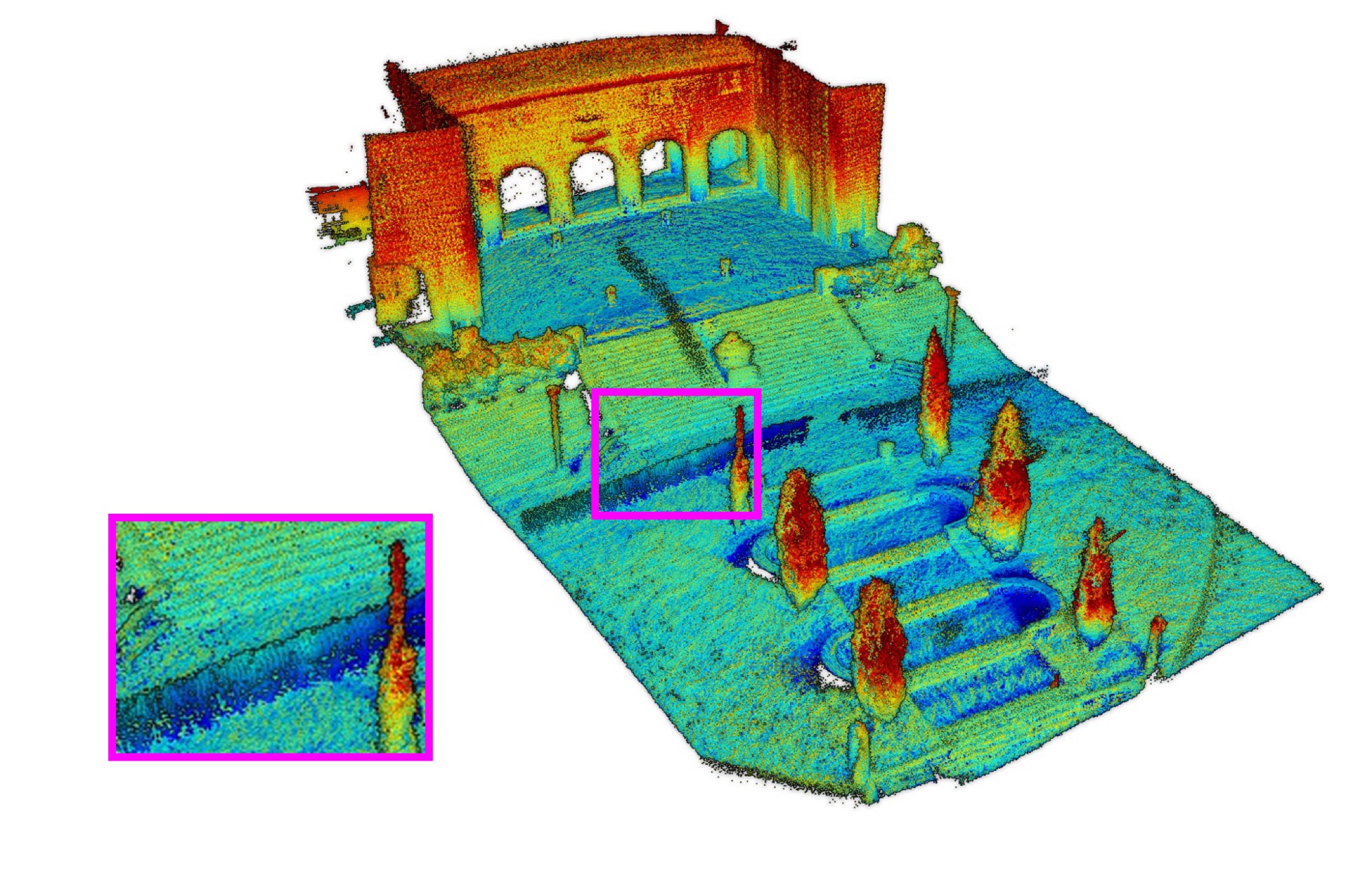}}
                                & \bpanel{\includegraphics[width=\bannerw,height=\bannerh,keepaspectratio,trim=84 20 41 12,clip]{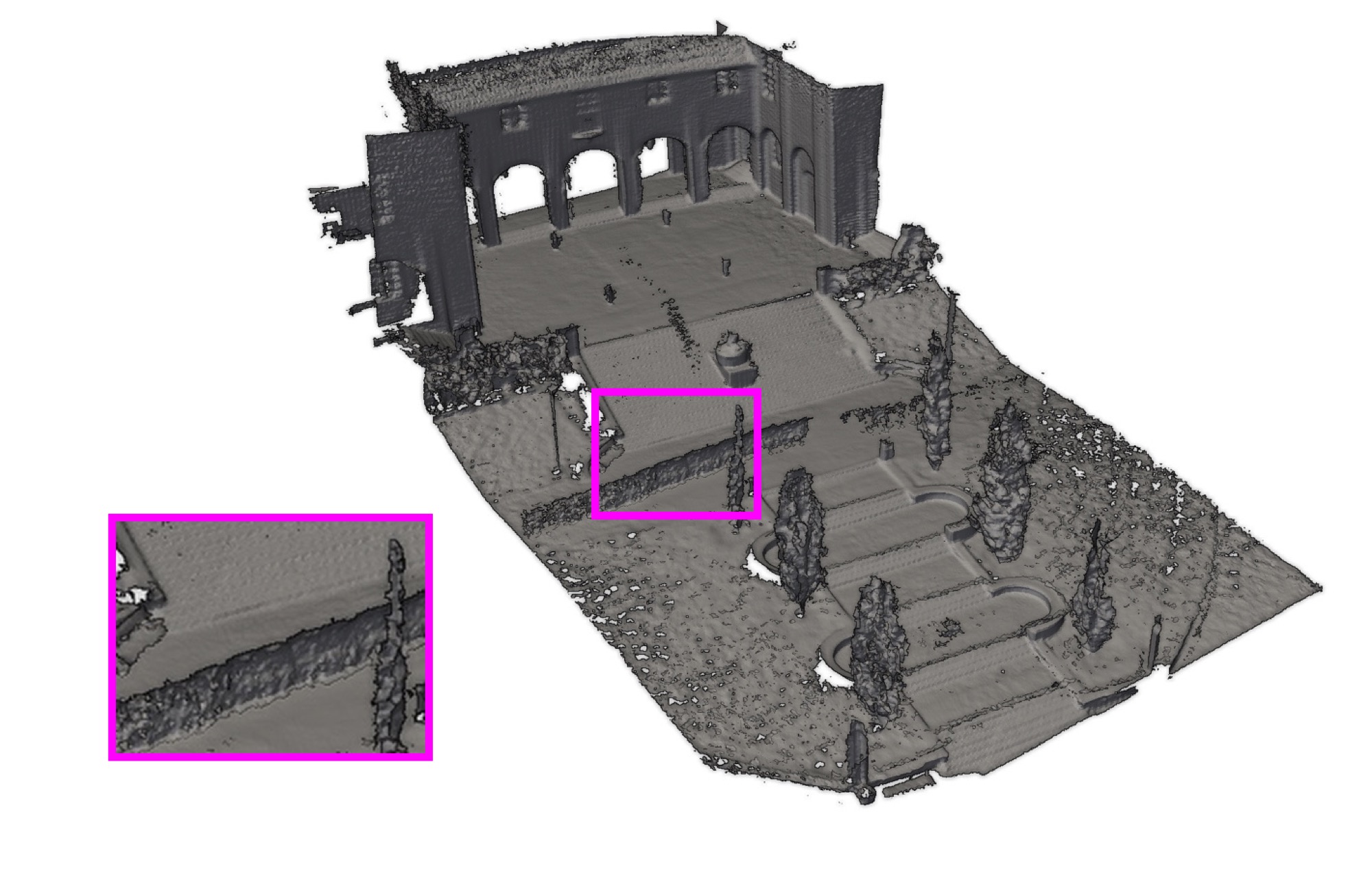}}
                                & \bpanel{\includegraphics[width=\bannerw,height=\bannerh,keepaspectratio,trim=84 20 41 12,clip]{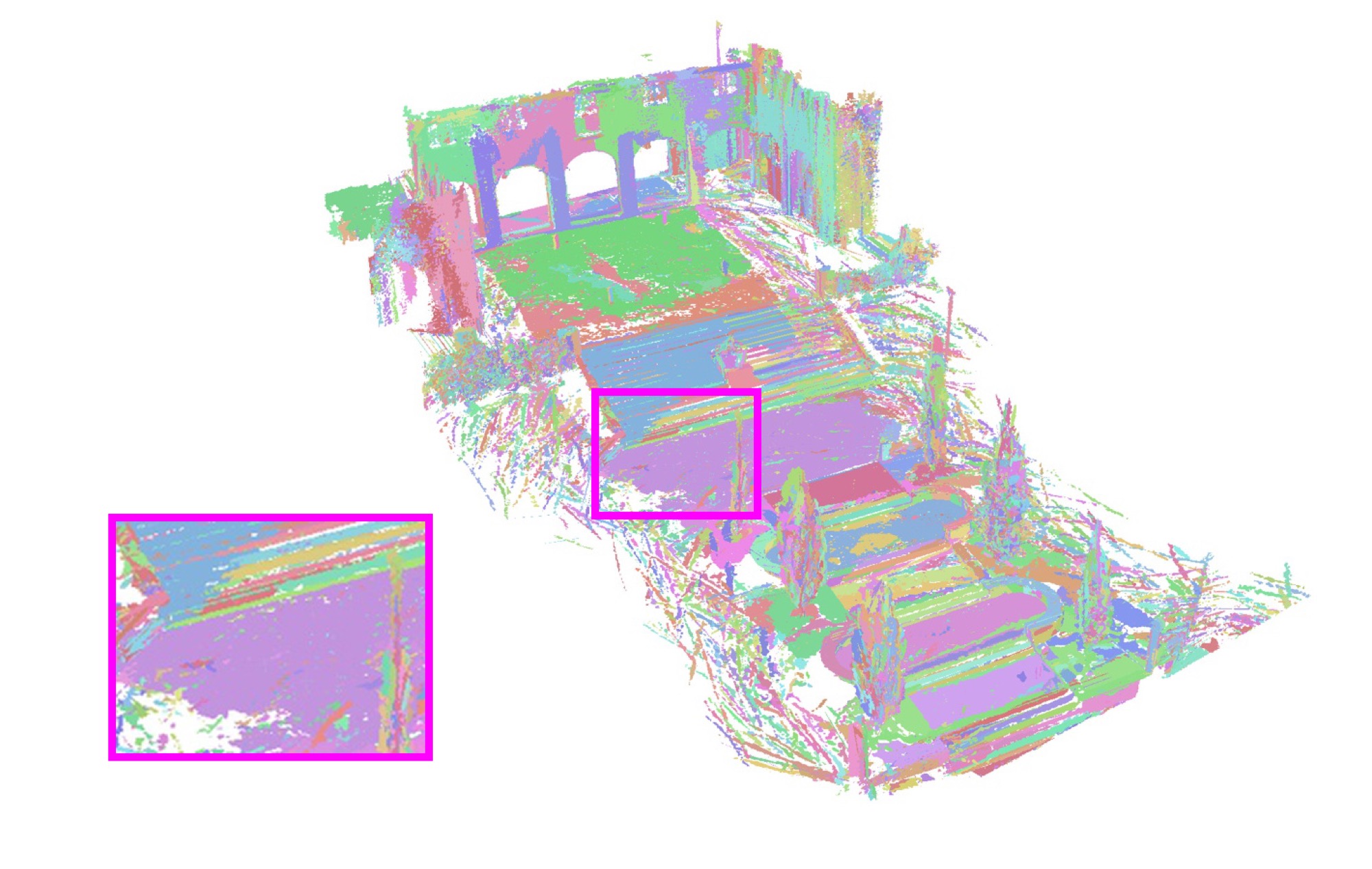}}
                                & \bpanel{\includegraphics[width=\bannerw,height=\bannerh,keepaspectratio,trim=84 20 41 12,clip]{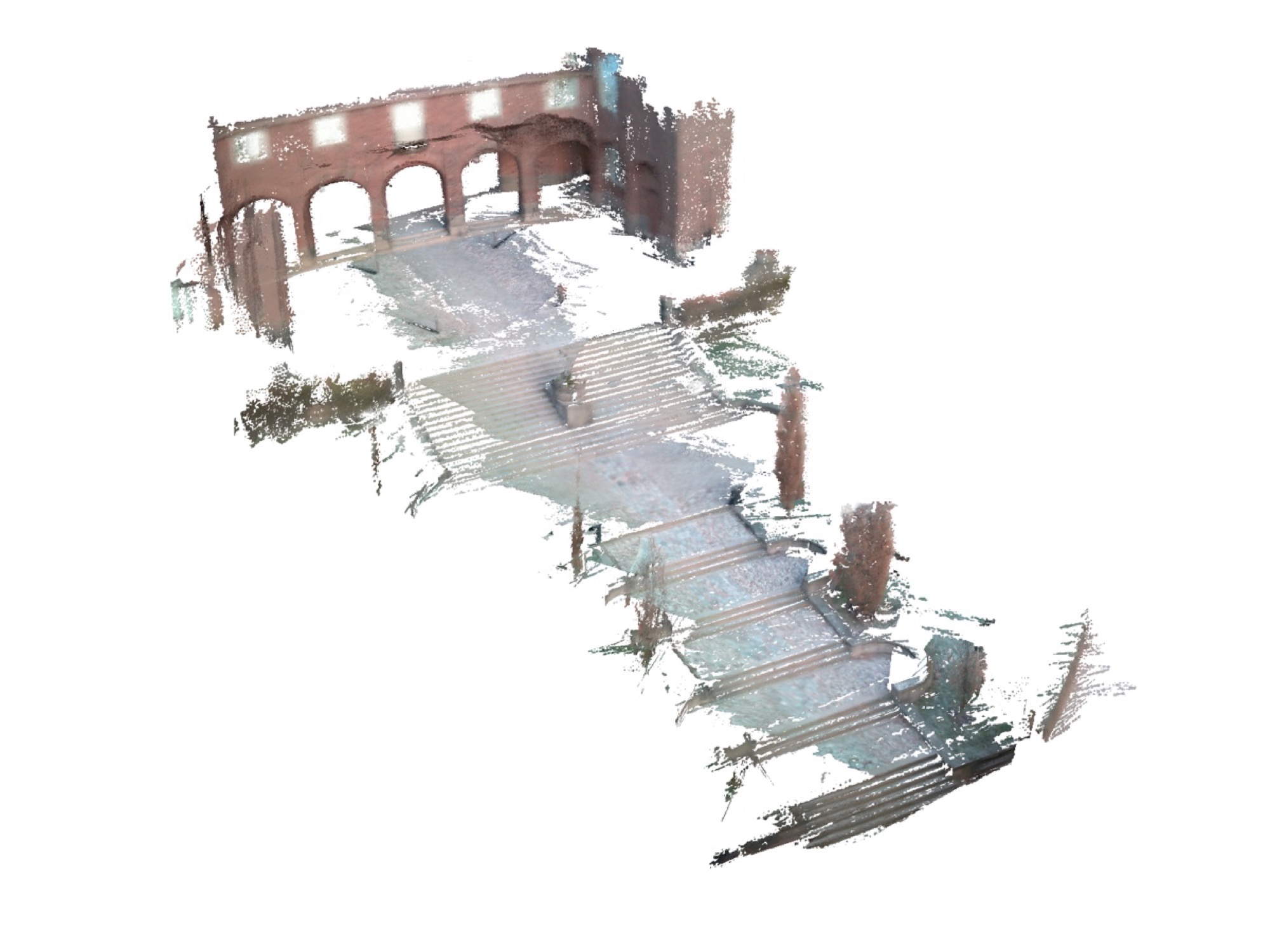}} \\
                \end{tabular}
                \caption{Qualitative reconstruction comparison across OSD (Keble03, Bodleian02) and MCD (kth\_day\_10). Each row shows: (a)~ground-truth tripod reference scan (not used during reconstruction); (b)~accumulated raw LiDAR scan coloured by beam ring index; (c)~volumetric mesh baseline (VDBFusion~\cite{VDBFusion}); (d)~OPP planar patches (ours); and (e)~textured OPP model (ours). Insets in row~3 (b--d) highlight walking pedestrians: dense motion trails visible in raw LiDAR scan (b) and integrated into the volumetric mesh (c) are cleanly carved away in OPP (d) via range-space transmission clearance (\refsec{sec:occupancy}). In textured OPP (e), patch areas left unobserved by MCD's narrower camera coverage remain white.}
                \label{fig:banner_long}
        \end{figure*}

                \reftab{tab:quantitative_results} reports quantitative comparisons across all sequences, and \reffig{fig:banner_long} illustrates qualitative reconstructions on a representative selection of sequences spanning both datasets.

                Across all six benchmark sequences, OPP consistently achieves the highest precision at \SI{5}{\centi\metre} and the lowest mean accuracy error, while matching or exceeding the best mode accuracy on four sequences (\reftab{tab:quantitative_results}). Free-space carving also naturally rejects dynamic obstacles (\reffig{fig:banner_long}, insets in row~3). While walking pedestrians leave dense motion trails in the raw LiDAR scans and the volumetric mesh, OPP's transmission term carves away these transient artifacts as subsequent rays observe the ground plane.

                The primary trade-off of our method lies in completeness and recall. At long ranges, beam divergence and sparse returns increase the in-plane uncertainty $\delta$. Rather than fitting false planes to noisy points or vegetation, OPP's gating and adaptive allocation suppress surface generation in uncertain regions. While this lowers recall on distant foliage (e.g., 28.2\% on kth\_day\_10 vs.~60.5\% for ImMesh), it ensures architectural boundaries remain cleanly defined without bridging across depth discontinuities.

                OPP also produces compact maps, achieving the smallest exported file size on Keble03 (\SI{30.2}{\mebi\byte}) and Bodleian02 (\SI{104.6}{\mebi\byte}), and ranking second on Observatory01 and kth\_day\_10. This efficiency stems directly from our representation: each supporting plane condenses thousands of coplanar points into a few geometric parameters, while 2D CSRBF kernels delineate surface boundaries without the heavy overhead of 3D grids or meshes.

\begin{figure}[ht]
                        \centering
                        \includegraphics[width=\columnwidth]{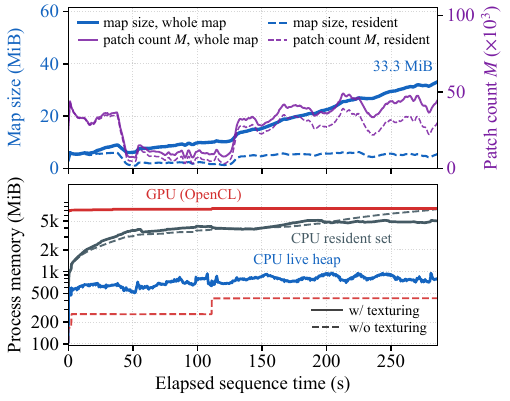}
                        \caption{Memory and map growth over time on Keble03 (\SI{2.1}{\second} rolling mean). \emph{Top}: map size and patch count $\npatch$, whole map (solid) vs.~resident (dashed). \emph{Bottom}: process memory with (solid) and without (dashed) texturing.}
                        \label{fig:memory_time}
                \end{figure}

                Beyond compact stored maps, OPP maintains a strictly bounded memory footprint during online operation (\reffig{fig:memory_time}). As newly explored facades expand the cumulative map on disk, disk-backed patch paging keeps resident RAM below \SI{7.6}{\mebi\byte} (median \SI{5.0}{\mebi\byte}), decoupling working memory from trajectory length. Similarly, texture projection offloads patch atlases to GPU memory (\SI{7.4}{\gibi\byte}), keeping the live CPU heap constant at $\sim$\SI{0.8}{\gibi\byte}.

\begin{figure}[t]
                        \centering
                        \includegraphics[width=\columnwidth]{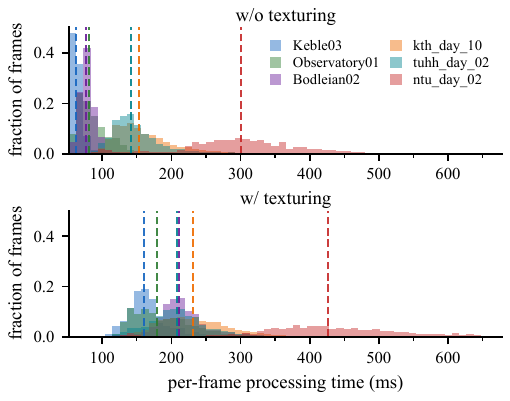}
                        \caption{Per-frame processing time distributions across the six benchmark sequences without (\emph{top}) and with (\emph{bottom}) texture projection. Histograms are normalised by frame count; dashed lines mark sequence medians.}
                        \label{fig:frame_time_hist}
                \end{figure}

                \reffig{fig:frame_time_hist} reports runtime distributions of OPP, with and without texture projection. On OSD, OPP runs comfortably within the \SI{10}{\hertz} sensor rate (median \SIrange{62}{81}{\milli\second} for $\sim$\num{30}k points per frame). On MCD, runtimes increase as input point count increases: \SIrange{142}{153}{\milli\second} for the OS1-64 sequences ($\sim$\num{57}k points per frame) and \SI{301}{\milli\second} for the denser OS1-128 sequence ($\sim$\num{94}k points per frame).
                Texture projection adds \SIrange{98}{134}{\milli\second} across OSD's three $1440\times1080$ cameras and \SIrange{66}{125}{\milli\second} for MCD's single $640\times480$ camera. Importantly, texturing latency is governed by A-buffer visibility sorting (\refsec{sec:texture}) rather than pixel count: doubling the number of active patches (\num{35}k to \num{71}k on MCD) increases texturing time by $1.9\times$, whereas a 15-fold increase in total image pixels adds only $\sim$\SI{40}{\milli\second}.

                In summary, OPP achieves real-time geometric reconstruction at the \SI{10}{\hertz} sensor rate on OSD, while disk-backed paging keeps resident RAM strictly bounded below \SI{8}{\mebi\byte} regardless of trajectory length. Online texturing adds moderate overhead governed by patch visibility, demonstrating that OPP achieves lightweight, high-fidelity architectural mapping on commodity hardware.

        \subsection{Ablation Studies}\label{sec:ablation}
                We evaluate the contribution of our two primary algorithmic components on Keble03 and Observatory01 (\reftab{tab:ablation_study}):
                (1) \emph{w/o EKF} replaces the range-space recursive plane update with standard isotropic PCA plane fitting.
                (2) \emph{w/o DP-GMM} replaces probabilistic Bayesian mixture assignment with hard Euclidean nearest-neighbour association and instantiating a new patch when a return falls outside a $3\sigma_{\expectedrange}$ distance gate.

                Disabling the range-space EKF consistently degrades accuracy and precision on both sequences (e.g., accuracy error rises from \SI{0.030}{\metre} to \SI{0.034}{\metre} on Keble03), demonstrating the systematic bias of isotropic estimators under anisotropic LiDAR range noise.

                Disabling the DP-GMM degrades accuracy while substantially inflating the stored map (e.g., from \SI{30.2}{\mebi\byte} to \SI{110.9}{\mebi\byte} on Keble03). A fixed distance threshold is too rigid for mature planes, initializing redundant duplicate patches, yet too soft for new planes, assigning ambiguous boundary returns which degrade reconstruction accuracy. The DP-GMM assignment is therefore essential to prevent map oversegmentation while preserving geometric precision.

                \begin{table}[t]
                        \centering
                        \caption{Ablation of system components, evaluated using the same metrics as in \reftab{tab:quantitative_results} (at \SI{5}{\centi\metre}). Completeness, recall, and F1 are greyed out as our design prioritises reconstruction accuracy and precision alongside map compactness. Bold indicates the best performance.}
                        \label{tab:ablation_study}
                        \footnotesize
                        \setlength{\tabcolsep}{4pt}
                        \resizebox{\columnwidth}{!}{
                        \begin{tabular}{l c c | c c c | c}
                                \hline
                                & & & \multicolumn{3}{c|}{\textbf{at \SI{5}{\centi\metre}}} & \\
                                \textbf{Config.} & \textbf{Mean Acc.} & \textbf{Mean Comp.} & \textbf{Prec.} & \textbf{Rec.} & \textbf{F1} & \textbf{Map Size} \\
                                & (m) $\downarrow$ & (m) $\downarrow$ & $\uparrow$ & $\uparrow$ & $\uparrow$ & (MiB) $\downarrow$ \\
                                \hline
                                \emph{Keble03}&&&&&& \\
                                w/o EKF & 0.034 & \textcolor{gray}{0.857} & 83.9\% & \textcolor{gray}{62.9\%} & \textcolor{gray}{71.9\%} & 37.8 \\
                                w/o DP-GMM & 0.033 & \textcolor{gray}{0.862} & 84.5\% & \textcolor{gray}{68.0\%} & \textcolor{gray}{75.3\%} & 110.9 \\
                                \textbf{Full OPP} & \textbf{0.030} & \textcolor{gray}{1.181} & \textbf{86.2\%} & \textcolor{gray}{62.2\%} & \textcolor{gray}{72.3\%} & \textbf{30.2} \\
                                \hline
                                \emph{Observatory01}&&&&&& \\
                                w/o EKF & 0.042 & \textcolor{gray}{0.449} & 76.5\% & \textcolor{gray}{66.7\%} & \textcolor{gray}{71.3\%} & 57.1 \\
                                w/o DP-GMM & 0.041 & \textcolor{gray}{0.394} & 77.5\% & \textcolor{gray}{71.2\%} & \textcolor{gray}{74.2\%} & 146.4 \\
                                \textbf{Full OPP} & \textbf{0.039} & \textcolor{gray}{0.506} & \textbf{79.2\%} & \textcolor{gray}{66.3\%} & \textcolor{gray}{72.2\%} & \textbf{50.5} \\
                                \hline
                        \end{tabular}
                        } %
                \end{table}

\section{CONCLUSIONS}

We presented OPP, an online incremental LiDAR surface reconstruction framework that couples supporting planes with continuous 2D occupancy fields. By formulating likelihoods, geometric updates, and ray clearance in range-space, OPP models anisotropic sensor noise and parameter uncertainties in closed form. Across six benchmark sequences, OPP achieves the highest precision and lowest accuracy error while producing stored maps among the most compact of the evaluated methods, providing cleanly parameterised surfaces suitable for architectural modelling.

In future work, we will perform quantitative evaluation on the textured reconstruction and investigate methods to join planar patch edges into closed architectural geometry.

\IEEEtriggeratref{14}
\IEEEtriggercmd{\enlargethispage{2\baselineskip}}
\bibliographystyle{IEEE/IEEEtran}
\bibliography{abrv,refs}

\end{document}